\documentclass[10pt,journal,compsoc,twocolumn]{IEEEtran}

\usepackage{amssymb}
\usepackage{bbding}
\usepackage{booktabs}
\usepackage{multirow}
\usepackage{flushend}
\usepackage{hyperref}
\usepackage{dsfont,amsfonts,amssymb,amsmath,color}
\usepackage{latexsym}
\usepackage{amsmath}
\usepackage{amsfonts}
\usepackage{graphicx}
\usepackage{color}
\usepackage{array}
\usepackage{algpseudocode}
\usepackage{makecell}
\usepackage{cases}
\usepackage{color}
\usepackage{amsthm}

\usepackage{soul}
\usepackage{tabularx}
\usepackage{makecell}
\usepackage{mwe}
\usepackage{subfig}
\usepackage{fixltx2e}

\usepackage[ruled,linesnumbered,vlined]{algorithm2e}
\usepackage{etoolbox}
\newlength\mylen

\makeatletter
\newcommand{\algorithmfootnote}[2][\footnotesize]{%
  \let\old@algocf@finish\@algocf@finish% Store algorithm finish macro
  \def\@algocf@finish{\old@algocf@finish% Update finish macro to insert "footnote"
    \leavevmode\rlap{\begin{minipage}{\linewidth}
    #1#2
    \end{minipage}}%
  }%
}
\makeatletter

\def\hlinewd#1{%
  \noalign{\ifnum0=`}\fi\hrule \@height #1 \futurelet
  \reserved@a\@xhline}
\makeatother

\usepackage{caption}
\begin{document}

\title{Uncertainty-guided Boundary Learning for Imbalanced Social Event Detection}

\author{Jiaqian Ren,
        Hao Peng,
        Lei Jiang,
        Zhiwei Liu,
        Jia Wu,
        Zhengtao Yu,
        Philip S. Yu,~\IEEEmembership{Fellow,~IEEE}
        % <-this % stops a space
\IEEEcompsocitemizethanks{
\IEEEcompsocthanksitem Jiaqian Ren and Lei Jiang are with the Institute of Information Engineering, Chinese Academy of Sciences, and the School of Cyber Security, University of Chinese Academy of Sciences, Beijing, China. E-mail: \{renjiaqian, jianglei\}@iie.ac.cn;
\IEEEcompsocthanksitem Hao Peng is with the School of Cyber Science and Technology, Beihang University, Beijing 1000191, China. E-mail: penghao@buaa.edu.cn;
\IEEEcompsocthanksitem Zhiwei Liu is with Salesforce AI Research, CA 94301, USA. E-mail: zhiweiliu@salesforce.com;
\IEEEcompsocthanksitem Jia Wu is with the School of Computing, Macquarie University, Sydney, Australia. E-mail: jia.wu@mq.edu.au;
\IEEEcompsocthanksitem Zhengtao Yu is with the Faculty of Information Engineering and Automation, and Yunnan Key Laboratory of Artificial Intelligence, Kunming University of Science and Technology, Kunming 650500, China. E-mail: ztyu@hotmail.com;
\IEEEcompsocthanksitem Philip S. Yu is with the Department of Computer Science, University of Illinois at Chicago, Chicago, IL 60607, USA. E-mail: psyu@.uic.edu.
}
\thanks{Manuscript received November 2022, major revised May 2023, major revised August 2023, accepted October 2023. (Corresponding author: Hao Peng.)}
}

% The paper headers
\markboth{}%
{Shell \MakeLowercase{\textit{et al.}}: Bare Demo of IEEEtran.cls for Computer Society Journals}

%==================================================================================
\IEEEtitleabstractindextext{%
\begin{abstract}
Real-world social events typically exhibit a severe class-imbalance distribution, which makes the trained detection model encounter a serious generalization challenge. 
Most studies solve this problem from the frequency perspective and emphasize the representation or classifier learning for tail classes. 
While in our observation, compared to the rarity of classes, the calibrated uncertainty estimated from well-trained evidential deep learning networks better reflects model performance. 
To this end, we propose a novel uncertainty-guided class imbalance learning framework - UCL$_{SED}$, and its variant - UCL-EC$_{SED}$, for imbalanced social event detection tasks. 
We aim to improve the overall model performance by enhancing model generalization to 
those uncertain classes. 
Considering performance degradation usually comes from misclassifying samples as their confusing neighboring classes, we focus on boundary learning in latent space and classifier learning with high-quality uncertainty estimation. 
First, we design a novel uncertainty-guided contrastive learning loss, namely UCL and its variant - UCL-EC, to manipulate distinguishable representation distribution for imbalanced data. 
During training, they force all classes, especially uncertain ones, to adaptively adjust a clear separable boundary in the feature space. 
Second, to obtain more robust and accurate class uncertainty, we combine the results of multi-view evidential classifiers via the Dempster-Shafer theory under the supervision of an additional calibration method. 
We conduct experiments on three severely imbalanced social event datasets including Events2012\_100, Events2018\_100, and CrisisLexT\_7. 
Our model significantly improves social event representation and classification tasks in almost all classes, especially those uncertain ones.

\end{abstract}

% Note that keywords are not normally used for peerreview papers.
\begin{IEEEkeywords}
Social event detection, evidential deep learning, demperster-shafer theory, imbalanced data
\end{IEEEkeywords}}

%==========================================================================================

% make the title area
\maketitle

\IEEEdisplaynontitleabstractindextext
% \IEEEdisplaynontitleabstractindextext has no effect when using
% compsoc or transmag under a non-conference mode.

\IEEEpeerreviewmaketitle

\IEEEraisesectionheading{\section{Introduction}\label{sec:introduction}}
Social event detection (SED) aims to correctly categorize the numerous social messages to detect the occurrences of events. 
Due to its wide application, recent years have witnessed lots of research on the detection methods~\cite{fedoryszak2019real,macedo2015context,li2014social}. 
However, few works investigate the severe data distribution imbalance problem in SED. 
Events have varying recognition difficulty levels because of the following two reasons. 
First, in the real-world scenario, event data typically exhibit a long-tail distribution with few head-dominant event classes and many low-frequent tail classes. 
Lacking sufficient training samples, the trained model's detection abilities for most events are data-sensitive, which means they are easily affected by per-class sample qualities. 
Second, some events may share semantically similar contexts with other events. 
This semantic-level overlapping issue also increases the complexity of event detection.

\begin{figure}
    \centering
    \includegraphics[width = 8.5cm]{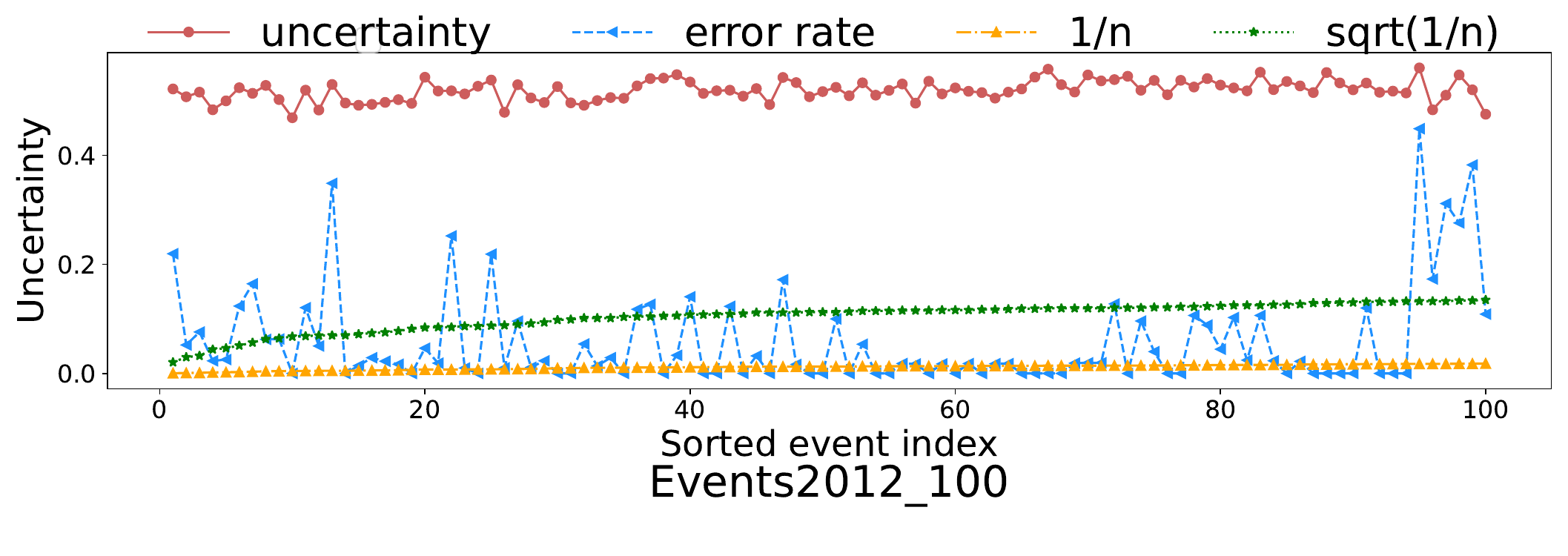}
    \caption{Statistics of the estimated per-class uncertainty, per-class error rate, the inverse of per-class number $1/n$ and the sqrt of $1/n$ on the training sets of Events2012\_100.%, Events2018\_100, and CrisisLexT\_7, respectively.
    }
    \label{fig_toy}
\end{figure}

Early approaches mainly focus on learning a balanced classifier to tackle the data imbalance issue. 
Two common strategies are: 
1) data re-sampling~\cite{more2016survey,buda2018systematic,pouyanfar2018dynamic}, whose core idea is to increase the numbers of samples in the tail classes or decrease the samples of those head classes; 
and 2) loss reweighting~\cite{cui2019class,wang2017learning,ren2018learning,tan2020equalization}, in which weights assigning to tail classes are larger. 
Though unbiased classifiers can be obtained by applying both aforementioned strategies, some recent works~\cite{kang2019decoupling,zhou2020bbn,ye2020identifying,jitkrittum2022elm} argue that they are unable to explicitly control the latent representation space, and therefore, are sub-optimal. 
Inspired by the intuition that qualitative features improve the classification, a recent line of work~\cite{kang2020exploring,wang2020understanding,wang2021contrastive,li2022targeted,zhu2022balanced} focuses on learning more separable representations for imbalanced data.
For example, authors in~\cite{wang2021contrastive} design a hybrid framework with a supervised contrastive learning branch for representation regularization and a classifier branch for bias eliminating. 
Later, some approaches~\cite{li2022targeted,zhu2022balanced} modify the original contrastive learning loss under the guidance of class frequency to further improve representation learning for imbalanced data. 
For example, BCL~\cite{zhu2022balanced} incorporates class-averaging and class-complement strategies to strengthen tail classes. 
DRO-LT~\cite{samuel2021distributional} learns high-quality representations based on distributional robustness optimization. 
% In essence, they adjust the class boundaries in latent space under the guidance of class frequency. 
However, as the difficulty level of event recognition is also affected by class overlapping, the class frequency may be an insufficient indicator to model performance. 
Our previous work~\cite{ren2022evidential} shows that evidential uncertainty estimated from well-trained EDL neural networks highly correlates with performance error. 
This correlation is shown in Fig.~\ref{fig_toy}. 
Compared to class frequency, the predicted uncertainty better indicates model generalization capacity. 
Therefore, in this paper, we explore a new direction toward learning balanced and separable representations under the guidance of uncertainty. 
The focus shifts from tail classes to uncertain classes. We aim to improve the overall model performance by improving learning for those uncertain ones.
To clarify, we also define uncertain event classes as events that are difficult to recognize due to the lack of labeled data or excessive noise from semantically similar events. 
Our goal is to propose a unified framework that combines the learning of uncertainty in the classification head with the uncertainty-guided boundary adjustment in the representation head.

To achieve this goal, in this work, we propose a novel uncertainty-guided class imbalance learning framework, namely UCL$_{SED}$, as well as its variant - UCL-EC$_{SED}$, for imbalanced social event detection tasks. It in essence works as a representation regularization technique guided by class-dependent uncertainty.
The core of our framework is to make adaptive and automatic boundary adjustments in the latent space. 
The blurry boundary gets separated by designing a loss that assigns larger margins in the latent space for those more uncertain classes.
Meanwhile, given that our assumption roots in high-quality uncertainty estimation, we also emphasize robust and accurate evidential classifier learning.  
Specifically, we design a novel uncertainty-guided contrastive learning loss (UCL and UCL-EC) to manipulate distinguishable representation distribution. To ensure robustness, in the classification head, the final uncertainty is calculated by combining multi-view results via Dempster's rule~\cite{sentz2002combination}. To ensure accuracy, an additional calibration method is utilized to prevent uncertain true predictions and certain false predictions.

We conduct extensive experiments on three imbalanced event datasets to evaluate our model.
Experimental results show our method achieves great results in all classes, especially in those hard (uncertain) classes. 
This demonstrates the superiority of our model. 
The code of this work is publicly available at GitHub\footnote{\url{https://github.com/RingBDStack/UCL\_SED}}.
\section{Related work}\label{sec:related_work}

\textbf{Social Event Detection.} 
Social event detection is a long-standing and challenging task. 
% Since an event appearing on a social platform involves social user networks and text streams, this task is much more complicated than pure social network mining and text classification tasks. 
% Also, text semantic mining in social networks is much more difficult for the following reasons. 
First, distinct from the formal texts that appeared on other occasions, text contents in social networks are often restricted to be pretty short and contain many informal expressions~\cite{peng2022reinforced}. 
These characteristics of social texts make the information extracted from original semantic text mining technologies far from satisfactory. 
Second, to depict an event, short social texts usually contain rich social network attributes~\cite{cui2021mvgan}, such as hashtags, timestamps, users, mentions, retweets, and so on. 
It is difficult to incorporate these heterogeneous attributes effectively. 
% Short informal text contents and rich heterogeneous network attributes cause the main challenges in social event detection.  
According to the utilized information, social event detection approaches can be roughly divided into three categories: content-based methods~\cite{zhao2011comparing,yan2015probabilistic,amiri2016short,wang2017neural,sahnoun2020event,morabia2019sedtwik}, attribute-based methods~\cite{xie2016topicsketch,xing2016hashtag,feng2015streamcube} and content-attribute-combining methods~\cite{zhou2014event,liu2020event,wang2016using,peng2019fine,cao2021knowledge,peng2021streaming,peng2022reinforced,cui2021mvgan,ren2022evidential}. 
As for content-based methods, a series of works make detection by analyzing text semantics. 
This type of method typically builds on text representation models such as Bag-of-words model~\cite{pradhan2019event}, Word2Vec~\cite{mikolov2013efficient} and Bert~\cite{devlin2018bert} or topic models like LDA~\cite{blei2003latent} to represent social texts. 
Because the text contents are short, which makes the captured information insufficient, some leverage multi-task technologies to extend the original knowledge. 
For example, authors in \cite{wang2017neural} utilize Deep Neural Networks to jointly make event detection and summarization. 
Some even incorporate information from external knowledge bases. 
As for the line of attribute-based methods, many studies make event detection by using important social attributes, such as hashtags~\cite{xing2016hashtag}, mention~\cite{guille2014mention,guille2015event}, retweet~\cite{chen2018social} and so on. 
This kind of work ignores the text semantics and thus is also insufficient. To grasp more comprehensive information, there is a trend toward content-attribute-combining methods.
They propose to integrate the content and multiple attributes with fusion or graph models. Due to their powerful expressiveness for graph data, in recent years, Graph Neural Networks have attracted lots of attention in the social event detection domain~\cite{ren2022known,peng2019fine,cao2021knowledge,peng2021streaming,peng2022reinforced,cui2021mvgan,ren2021transferring,ren2022evidential,ren2022known}.
These GNN-based detection methods build heterogeneous information networks to represent social event data. 
Various attributes in social networks effectively complement each other and play an independent role in text semantic propagation and aggregation.
For example, KPGNN~\cite{cao2021knowledge} utilizes users, keywords, and entity attributes to construct an event message graph and then leverages inductive Graph Attention Networks (GAT) to learn message representations. 
PP-GCN~\cite{peng2019fine} utilizes multiple attributes by designing sophisticated meta-paths and then uses Graph Convolutional Networks (GCN) to obtain representations. Later, some works leverage multi-view learning strategies to further strengthen the feature learning process. 
MVGAN~\cite{cui2021mvgan} learns message features from both the semantic and temporal views, then proposes an attention mechanism to fuse them together. 
ETGNN~\cite{ren2022evidential} instead learns representations from \textit{co-hashtag}, \textit{co-entity} and \textit{co-user} views. 
Dempster-Shafer Theory extracts shared beliefs to resist noise and make the final decision more robust. 
Our work builds on ETGNN and makes a modification in the representation learning process to adapt to imbalanced data.
Besides, we also add an uncertainty calibration method to ensure the accuracy of the estimated uncertainty.

\textbf{Long-tail Recognition.} 
Real-world data such as events, images and objects usually follow a long-tail distribution, which makes the trained model generalize badly.
Early solutions tend to solve this problem from two perspectives: 
(1) modifying the training data through some data re-sampling strategies~\cite{more2016survey,buda2018systematic,pouyanfar2018dynamic,chawla2002smote,han2005borderline,beery2020synthetic,kim2020m2m, Drummond2003} and (2) modifying the loss function with re-weighting strategy~\cite{cui2019class,wang2017learning,ren2018learning,tan2020equalization,ryou2019anchor} or margin modification strategy~\cite{menon2020long,khan2019striking,cao2019learning}. 
Re-sampling approaches mainly consist of three popular techniques, including over-sampling tail classes by sample copying~\cite{chawla2002smote,han2005borderline}, generating augmented samples to supplement tail classes~\cite{pouyanfar2018dynamic,beery2020synthetic,kim2020m2m} and under-sampling head classes by discarding part of data~\cite{Drummond2003}. 
Despite the good results, these three techniques face tail-class overfitting, expensive cost, and model generalization problems, respectively~\cite{chawla2002smote,more2016survey}. 
Loss re-weighting approaches tailor the loss function based on up-weighting the samples in tail classes and down-weighting those in head classes.
For instance, works such as~\cite{he2009learning,cui2019class} re-weight the loss functions by the inverse of class frequencies. In the work~\cite{ren2018learning}, a held-out evaluation set is utilized to optimize the weights to samples. 
Authors in \cite{ryou2019anchor} leverage the difficulty level of sample prediction, measured by the confidence score gap, to rescale the cross-entropy loss.
Also, there is an alternative strategy to manipulate the classification loss margins. 
Some works~\cite{menon2020long,cao2019learning} proposed encouraging larger logit margins for rare classes and decreasing margins for head classes.

All the aforementioned methods focus on learning a balanced classifier. While recently, some works~\cite{kang2019decoupling,zhou2020bbn,jitkrittum2022elm} explore decoupling the original classification learning into two separate stages including representation learning and classifier learning. 
According to their observation, the learned representation with a balanced classifier (trained by re-sampling and re-weighting methods) is sub-optimal.
Based on this observation, more and more works turn to learn better representations from imbalanced data to improve classification performance~\cite{yang2020rethinking,wang2021contrastive,zhu2022balanced,samuel2021distributional,li2022targeted}. 
Inspired by the great promise of contrastive learning~\cite{chen2020simple,khosla2020supervised} in obtaining distinguishable representations, researchers have investigated the potential of leveraging contrastive learning loss to manipulate further performance gain. 
SSP~\cite{yang2020rethinking} leverages self-supervised and semi-supervised contrastive learning to boost long-tailed learning tasks. 
Authors in \cite{wang2021contrastive} design a hybrid framework with a supervised contrastive learning branch for better representation and a classifier branch for bias elimination. 
Sharing a similar framework, the work~\cite{zhu2022balanced} introduces a novel BCL loss in the representation learning branch to deal with the domination of head classes. 
Based on distributional robustness optimization, DRO-LT~\cite{samuel2021distributional} explicitly seeks to improve the quality of representations for tail classes.
Authors in \cite{li2022targeted} instead propose targeted supervised contrastive learning with a set of pre-defined feature distributions. 
Most of these works modify the contrastive learning loss under frequency guidance. 
However, according to our observation, frequency is a worse indicator of model performance than uncertainty. Our work leverages uncertainty to adaptively adjust class boundary learning in the latent space. 
A similar work to ours is \cite{khan2019striking}, which links class imbalance problems with Bayesian uncertainty estimates. 
However, it indirectly models uncertainty through network weights, which is inefficient. 
Besides, it utilizes uncertainty to adjust logit margins instead of representation margins. 
\section{Preliminary}\label{sec:preliminary}
\subsection{Classification Task}
The aim of the classification task is to learn a complicated mapping function from an input space $\mathcal{X}$ to a target space $\mathcal{Y}=\{1, 2, ..., C\}$. 
Generally, the mapping function is composed of two parts: an encoder model $f$ which maps the input to a latent space $\mathcal{Z}\in \mathbb{R}^h$, and a classifier $g$ which maps the latent space $\mathcal{Z}$ to the target space $\mathcal{Y}$. 
In this work, we leverage a modified contrastive loss to adjust the latent space $\mathcal{Z}$. We tend to improve the final classification performance by making the learned representations more distinguishable.

\subsection{Temporal-aware GNN encoder}\label{sec_temporal}
Graph neural networks are proposed for representation learning on graph data~\cite{kipf2016semi,hamilton2017inductive}. 
For each node on the graph, a GNN encoder iteratively updates its representation by combining information from its one-hop neighbors. 
In this way, the learned representation contains graph structural and node attribute information and is more comprehensive. 
Typically, a GNN encoder layer comprises two types of operation: feature transformation operation and feature aggregation operation. 
Suppose the node representation of index $i$ in the $(l-1)-th$ layer is denoted as $\mathbf{h}_i^{l-1}$, its updated representation in the next layer is computed as follows:
\begin{equation}
    \boldsymbol{h}_{i}^{(l)} \leftarrow 
    \sigma\left(
     \underset{\forall j \in \mathcal{N}\left(i\right)}{\operatorname{Aggregator}}\left(\text { Transformation }\left(\boldsymbol{h}_{j}^{(l-1)}\right)\right)\right),
\end{equation}
where $\mathcal{N}\left(i\right)$ represents the set of neighbor indices of node $i$. 
Aggregator and Transformation are designed differently in different GNN models. 
Since temporal information is important in indicating events, we adopt the temporal-aware GNN aggregator in our previous work \cite{ren2022evidential} to incorporate temporal information into graph representation learning. 
As for the transformation operation, we use the simple linear trainable transformation.
The specific layer-wise propagation becomes:
\begin{equation}\label{eq_temGnn}
    \mathbf{h}_{i}^{l} \leftarrow \sigma\left(\sum_{j \in \mathcal{N}(i)} a_{ij} \mathbf{W h}_{j}^{l-1}\right).
\end{equation}
Here $\mathbf{W}$ denotes the transformation matrix learned during training. $\sigma(\cdot)$ is an activation function. Attention weight $a_{ij}$ measures the temporal approximation between message from node $i$ and message from node $j$ and is computed as follows:
\begin{equation}
a_{ij} = \frac{e^{-fc(h_i^l)\cdot|t_j-t_i|}}{\sum_{j\in\mathcal{N}(i)}e^{-fc(h_i^l)\cdot|t_j-t_i|}},
\end{equation}
where $t_i$ and $t_j$ are the publishing time of message $i$ and message $j$.
$|t_j-t_i|$ is the corresponding time interval that can be counted in days, hours, minutes, etc. 
Here in this paper, we counted in days.
$fc(\cdot)$ represents a fully connection layer. More details can be seen in \cite{ren2022evidential}.

\begin{figure*}
    \centering
    \includegraphics[width = 16.5cm]{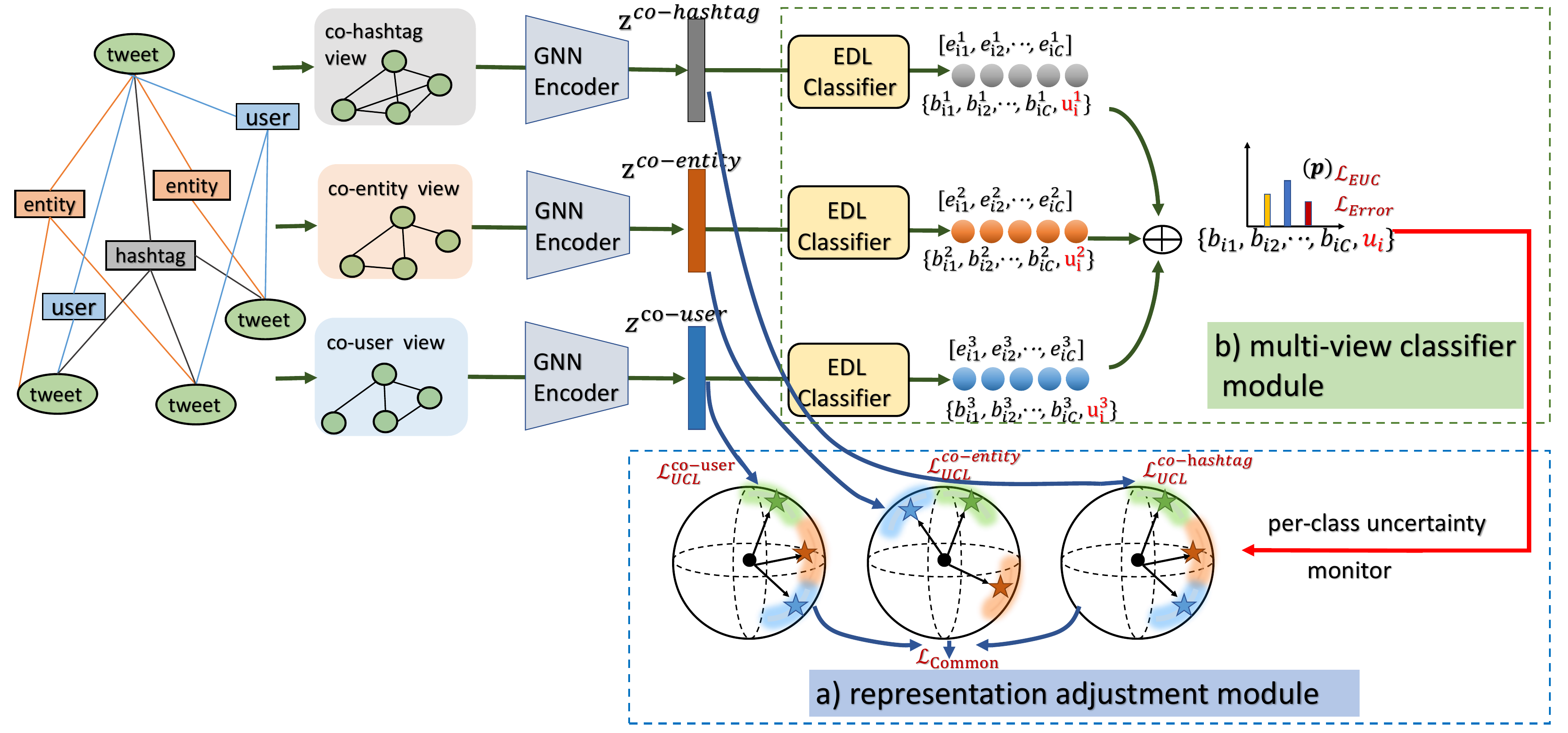}
    \caption{The architecture of the proposed uncertainty-guided class imbalance learning framework (UCL$_{SED}$ and UCL-EC$_{SED}$). The whole framework contains two modules: a) representation adjustment module, in which larger margins are assigned to more uncertain classes to ensure class separability; b) multi-view classifier module, in which multi-view results are combined via Dempster-Shafer theory with an additional calibration method to ensure robustness and accuracy.}
    \label{fig_model}
\end{figure*}

\subsection{Contrastive Learning}\label{sec_psc}
Contrastive learning imposes geometric constraints on the sample representations to regulate the model. 
It follows a simple principle of pulling the samples from the same class together and pushing the samples from different classes apart. 
We here introduce some variants of supervised contrastive learning loss which will facilitate the understanding of our later modification.

\textbf{Supervised contrastive loss~\cite{khosla2020supervised}}. 
Supervised contrastive loss (SCL) utilizes label information to find samples within the same class as the positive ones. 
Formally, in a batch $B$, for an instance $\mathbf{x}_i$ whose representation learnt by encoder $f$ is $\mathbf{z}_i$,
supervised contrastive loss is written as:
\begin{equation}
    \mathcal{L}_{SCL}(\mathbf{z}_i) =-\frac{1}{|\{x_i^+\}|} \sum_{j \in \{x_i^+\} } \log \frac{\exp \left(S(\mathbf{z}_{i} , \mathbf{z}_{j}) / \tau\right)}{\sum_{k \in B \backslash\{i\}} \exp \left(S(\mathbf{z}_{i} , \mathbf{z}_{k}) / \tau\right)},
\end{equation}
where $\{x_i^+\}$ denotes a subset of B that contains all samples within the same class as $x_i$. 
$|\{x_i^+\}|$ is the number of all the positive samples in the batch $B$. 
$\tau$ is a temperature parameter. $S$ denotes a similarity metric function where cosine similarity is often selected. 
Because cosine similarity removes the effect of feature length and emphasizes angle information, which further facilitates linear classification thus making the training process more stable. 
To sum up, SCL maximizes agreement between the anchor and all positive samples by contrasting against samples from other classes. 
While simple and effective, this loss faces memory issues~\cite {khosla2020supervised}.

\textbf{Prototypical supervised contrastive loss~\cite{wang2021contrastive}}. 
Prototypical supervised contrastive (PSC) loss replaces specific positive and negative samples with prototypes to tackle the memory issue. 
In PSC, each sample is pulled close to the prototype of its class and pushed away from prototypes of other classes. 
Formally, suppose there is a sample $x_i$ (representation is $z_i$) whose label is $y_i$, the PSC loss function can be expressed as follows:
\begin{equation}
    \mathcal{L}_{PSC}\left(\mathbf{z}_{i}\right)=-\log \frac{\exp \left(S(\mathbf{z}_{i} \cdot \mathbf{p}_{y_{i}}) / \tau\right)}{\sum_{c=1}^{\mathcal{C}} \exp \left(S(\mathbf{z}_{i} \cdot \mathbf{p}_{c}) / \tau\right)},
\end{equation}
where $\mathbf{p}_{y_{i}}$ is the prototype representation of class $y_i$ that sample $x_i$ belongs to. The prototype representations are learned during training. 
\section{The Proposed Model}\label{sec:model}
\subsection{Overall Framework}
We start with an overview of our uncertainty-guided class imbalance learning framework UCL$_{SED}$ and UCL-EC$_{SED}$ and provide the details in subsequent sections.

The overall framework is demonstrated in Fig.~\ref{fig_model}, built on our previous work - ETGNN~\cite{ren2022evidential}.
The aim of this paper is to enhance imbalanced social event detection by learning better representations such that the class boundaries are well separated. 
With the observation that evidential uncertainty well reflects model performance and the assumption that model performance is highly correlated with the representation distribution, the key idea becomes using the predicted evidential uncertainty from the classifier to monitor and adjust the representation distribution status during the training process.

The detailed training process of UCL$_{SED}$ is depicted in Algo.~\ref{Alg_model}. 
Following \cite{ren2022evidential}, we first construct three view-specific message graphs (\textit{co-hashtag}, \textit{co-entity} and \textit{co-user}) by simply connecting messages sharing the same corresponding element together. 
Then we utilize a temporal-aware GNN encoder (also introduced in Sec.~\ref{sec_temporal}) to obtain the message representations. 
The next steps are this paper's two key modules, which will be roughly introduced in the next paragraph and detailed in the subsequent sections.
Note that UCL-EC$_{SED}$ is similar to UCL$_{SED}$ but different in acquiring the prototypes. 
Considering the page limit, we didn't show the detailed algorithm of UCL-EC$_{SED}$. 
But readers can refer to Sec.~\ref{sec_ucl-ce} to see the concrete difference.

Two key modules in the framework are the representation adjustment module and the multi-view classifier module. 
In the representation adjustment module, to ensure class separability, we propose a novel uncertainty-guided contrastive learning loss (i.e. UCL and UCL-EC) to assign larger inter-class margins in the latent space for those uncertain classes. 
In the multi-view classifier module, we use Dempster-Shafer theory to combine the results from the three single views. 
Meanwhile, considering uncertainty plays an important role in representation adjustment and classification making, an additional calibration constraint is added to get better and more robust uncertainty estimations and class predictions. 
Generally, in our work, representation, and classification learning are combined closely and mutually promote each other. 
On one hand, the result from classification learning works as an effective indicator to reflect the current representation distribution status. 
The representations become more distinguishable by setting larger margins for more uncertain classes. 
On the other hand, in line with the property of intra-class compactness and inter-class separability, the adjusted representation further facilitates classification learning. 

\makeatletter
\patchcmd{\@algocf@start}
{-1.5em}
{0pt}
{}{}
\makeatother
\begin{algorithm}[h]
% \small
\KwIn{Imbalanced event dataset $\mathcal{X}$ with corresponding labels $\mathcal{Y}=\{1,2,...,C\}$, maximal epoch number $E$, views $v\in\{\textit{co-hashtag, co-entity, co-user}\}$, the number of GNN layers $L$, and the number of mini-batches $B$}
\KwOut{Parameters of the GNN encoder model $f(\theta)$, view-specific prototypes $p^v_c, c\in\{1,2,...,C\}$, parameters of the view-specific classifiers $g^v(\theta)$}
\For{$v\in\{\textit{co-hashtag, co-entity, co-user}\}$}{construct view-specific message graph $G^v$ as \cite{ren2022evidential}}
Initialize the parameters $f(\theta)$ and $g^v(\theta)$, initialize the view-specific prototypes $p^v_c, c\in\{1,2,...,C\}$, initialize the per-class uncertainty $[u_1, u_2, ..., u_C]$ as $[1-\epsilon,1-\epsilon,...,1-\epsilon]$, where $\epsilon$ is a small value.\\
\For{$e = 1, 2, ..., E$}{
         \For{$b = 1, 2, ..., B$}
         {
             Sample a mini-batch of messages $\{m_b\}$\\
             \For{$v\in\{\textit{co-hashtag, co-entity, co-user}\}$}
                    {%Obtain the sub-graph $G^v_b$\\
                    \For{$l = 1, 2, ..., L$}
                    {Obtain $h^{v(l)}_i, i \forall \{m_b\}$ via Eq.~\ref{eq_temGnn} 
                    }
                    $z^v_i\gets h^{v(L)}_i$, $i \forall \{m_b\}$\\
                    Calculate $\mathcal{L}_{UCL}^v$ via Eq.~\ref{eq_ucl}\\
                    $[e^v_{i1}, e^v_{i2}, ..., e^v_{iC}] \gets$ EDL classifier $g^v$\\
                    Obtain $[b^v_{i1}, b^v_{i2}, ..., b^v_{iC}, u_i^v]$ via Eq.~\ref{equ_edl}}
            Obtain $[b_{i1}, b_{i2}, ..., b_{iC}, u_i] $ via Eq.~\ref{eq_dst}\\
            Calculate $\mathcal{L}_{Error}$ via Eq.~\ref{eq_error}\\
            Calculate $\mathcal{L}_{EUC}$ via Eq.~\ref{eq_euc}\\
            Calculate $\mathcal{L}_{Common}$ via Eq.~\ref{eq_common}\\
            Calculate $\mathcal{L}_{Total}$ via Eq.~\ref{eq_total}\\
            Update $f(\theta)$, $g^v(\theta)$ and prototypes $p^v_c$\\
            }
            Update per-class uncertainty $[u_1, u_2, ..., u_C]$
        }

\caption{\textbf{Uncertainty-guided class imbalance learning framework (UCL$_{SED}$) for imbalanced social event detection.
}}
\label{Alg_model}
\end{algorithm}

\subsection{Representation Adjustment Module}

\subsubsection{Uncertainty-Guided Contrastive Learning Loss (UCL)}\label{sec_psc_ucl}
In this section, we introduce UCL in detail, an extension of prototypical supervised contrastive loss. 
The key difference lies in the setting of the class margin. 

To facilitate understanding, we rewrite PSC loss as follows:
\begin{equation}
\begin{aligned}
    \mathcal{L}_{PSC}\left(\mathbf{z}_{i}\right) &= \log \left[1+\sum_{c=1, c \neq y_{i}}^{\mathcal{C}} e^{  \Delta_{y_{i}c}+ S(\mathbf{z}_{i},\mathbf{p}_c)-S(\mathbf{z}_{i},\mathbf{p}_{y_{i}})}\right],\\\Delta_{y_{i}c}&=0.
\end{aligned}
\end{equation}
For simplicity, we remove the temperature parameter $\tau$. 
With the parameter that controls class margin being set to constant zero, PSC loss has a weak capacity to learn separable representation distribution for imbalanced data. 
While in practice, events are diverse and complex and have highly imbalanced frequencies. 
All these properties make different events face varying difficulty levels to be well-represented and correctly classified. 
Concretely, if an event is hard to be distinguished by a classification model (i.e., the classification model is uncertain about its prediction), there is a high probability that this event faces a blurry boundary in the latent space and thus is easy to be misclassified. 
Therefore, an adaptive and flexible regularization loss that sets proper class-dependent margins is highly needed for better representation distribution for imbalanced event data. 
For events that are hard to distinguish, intuitively, we should enlarge the margins of other events towards them. 
In other words, the modified loss should push the distribution of other events away from them to prevent misclassification. 
To achieve that, we modify the original PSC loss with a tunable margin controlled by uncertainty, turning into UCL as follows:
\begin{equation}\label{eq_ucl}
     \mathcal{L}_{UCL}\left(\mathbf{z}_{i}\right) = \log \left[1+\sum_{c=1, c \neq y_{i}}^{\mathcal{C}} e^{ \beta u_{y_{i}}+ S(\mathbf{z}_{i},\mathbf{p}_c)-S(\mathbf{z}_{i},\mathbf{p}_{y_{i}})}\right],
\end{equation}
where $u_{y_i}$ represents the uncertainty value of class $y_{i}$ and works as an effective indicator to reflect its current representation distribution status. 
$\beta$ is a positive hyperparameter. 
The computation of $u_{y_i}$ will be introduced in Sec.~\ref{sec_classifier}. 

\textbf{Comparison with the original PSC loss.} 
The modified UCL added with an additional positive value encourages a larger margin between prototypes. 
Therefore, it simultaneously enhances the intra-class compactness and inter-class discrepancy. 
Moreover, this loss can be considered as a soft approximation to $\max(0, \beta u_{y_{i}}+ S(\mathbf{z}_{i},\mathbf{p}_c)-S(\mathbf{z}_{i},\mathbf{p}_{y_{i}}) ), c=\arg\max_{c \neq y_{i}}(S(\mathbf{z}_{i},\mathbf{p}_c))$.
With the per-class margin being set to be positively related to class uncertainty, this loss is class-dependent. Meanwhile, the existence of the extra penalty $ \beta u_{y_{i}}$ forces even larger margins between $S(\mathbf{z}_{i},\mathbf{p}_{y_{i}})$ and $S(\mathbf{z}_{i},\mathbf{p}_c)$. Thus, events with large uncertainty are pushed more away from other classes to avoid class overlapping. 
In this way, the representation boundary of each class can be adjusted automatically and properly.

\subsubsection{Uncertainty-Guided Contrastive Learning Loss With Estimated Centroids (UCL-EC)}\label{sec_ucl-ce}
\textbf{A variant of UCL with estimated centroids (UCL-EC).} 
As seen in Algo.~\ref{Alg_model}, the prototypes in the UCL are learned during training and updated in each batch.
Considering samples in different batches vary a lot, the learned prototypes may fluctuate greatly in different batches. 
This training instability problem is particularly severe for imbalanced data, where most classes are minority classes and are likely to be sampled in different batches.
We replaced the prototypes with the global class centroids calculated beyond batch data to stabilize the training process. 
However, calculating class centroids in the full dataset costs time and computation resources. 
To make a trade-off between training stability and resource consumption, instead of updating centroids after the training of each batch, we estimate the centroids at the beginning of every epoch and keep them fixed in memory for the duration of the whole epoch.

\subsection{Multi-View Classifier Module}\label{sec_classifier}
\subsubsection{Single-View Uncertainty From EDL}\label{sec_edl}
There are two kinds of uncertainty: epistemic uncertainty, also called model uncertainty, results from limited knowledge; aleatoric uncertainty, also named data uncertainty, is the noise inherent from class overlap. 
For instance, samples distributed in the blurry class boundary have high aleatoric uncertainty. 
This work uses the measured class aleatoric uncertainty to monitor the class representation status. 
Intuitively, we assume a class with high aleatoric uncertainty is not well represented and should be emphasized in the representation adjustment module. 
In recent years, evidential deep learning (EDL) has been proposed to estimate aleatoric uncertainties by directly estimating parameters of the predictive posterior based on the output of the deep neural networks ~\cite{sensoy2018evidential}. 

\textbf{EDL for single-view event detection.} 
Under the framework of Subjective Logic and Dempster-Shafer theory~\cite{dempster2008upper}, EDL provides a principled way to jointly model high-order probabilities for a prediction and model uncertainty for the overall decision. 
Specifically, it assumes a Dirichlet distribution as the conjugate before the Multinomial distribution to represent the density of class probability assignment. 
The belief mass assignment to each event class and the overall uncertainty mass are determined over the Dirichlet distribution, and the Dirichlet parameters are induced from the collected evidence learned by the neural network. 

Formally, for each single view v, where $v\in\{\textit{co-hashtag, co-entity, co-user}\}$, suppose there are C mutually exclusive events, the Dirichlet distribution of the $i-th$ sample $\mathbf{\alpha_i}^v = [\alpha^v_{i1}, \alpha^v_{i2}, ..., \alpha^v_{iC}]$ is induced from the evidence $\mathbf{e_i}^v = [e^v_{i1}, e^v_{i2}, ..., e^v_{iC}]$ collected from the data with the relation $\alpha_{ic}^v = e_{ic}^v+1$, $c\in\{1, 2, ..., C\}$. 
The belief mass assignment to each event, as well as the overall uncertainty mass, is computed as follows:
\begin{equation}\label{equ_edl}
b_{ic}^v=\frac{e^v_{ic}}{S_i^v}, u_i^v=\frac{C}{S_i^v},
\end{equation}
where $S_i^v=\sum_{c=1}^C e_{ic}^v+1=\sum_{c=1}^C\alpha_{ic}^v$ is referred to as the Dirichlet strength. Obviously, more evidence ensures less uncertainty.

\subsubsection{Multi-View Uncertainty Via DST}\label{sec_dst}
After getting evidence-based single-view opinions, to ensure a more robust final result, we combine them together via Dempster-Shafer theory. 
The combination rule, also known as Dempster’s rule, strongly emphasizes the agreement between multiple views and extracts their common shared beliefs as the final judgment. 
Specifically, for the $i-th$ sample, we need to combine three independent sets of mass assignments $M_i^v=\{\{b^v_{ic}\},u_i^v\}$, where $v\in\{\textit{co-hashtag, co-entity, co-user}\}$. 
Here we utilize $\oplus$ to denote the dempster's combination rule in combining two independent views. 
The detailed calculation is as follows:
\begin{equation}\label{eq_dst}
    \begin{aligned}
    M_i &= M_i^{v_1}\oplus M_i^{v_2},\\
    b_{ic}&=\frac{1}{1-T_i}\left(b_{ic}^{v_1} b_{ic}^{v_2}+b_{ic}^{v_1} u_i^{v_2}+b_{ic}^{v_2} u_i^{v_1}\right), u_i=\frac{1}{1-T_i} u_i^{v_1} u_i^{v_2},\\
    T_i&=\sum_{j\neq k}b_{ij}^{v_1}b_{ik}^{v_2}.
    \end{aligned}
\end{equation}
The combination rule can be further extended to the multi-view case.
As the case in this paper, we have three sets of masses (i.e., masses learned under the three single views: \textit{co-hashtag}, \textit{co-entity} and \textit{co-user}). The final results can be obtained sequentially as follows:
\begin{equation}
 M_i = M_i^{v_1}\oplus M_i^{v_2}\oplus M_i^{v_3},
\end{equation}
where $v_1, v_2$ and $v_3$ correspond to \textit{co-hashtag}, \textit{co-entity} and \textit{co-user} views, respectively.

The above procedure describes the calculation of the uncertainty of each sample in detail. 
As for the uncertainty value of each class, we use the average uncertainty values of all the training samples within that class as its class uncertainty.

\subsubsection{Uncertainty Calibration Method}
Though the uncertainty can be modelled directly with EDL and DST, it may not be well calibrated~\cite{bao2021evidential}. 
Considering the important role the estimated uncertainty plays, we need it to be as accurate as possible to reflect the status of representation learning. 
We adopt an uncertainty calibration method to build the correct relationship between accuracy and uncertainty. 
This is also inspired by previous calibration studies~\cite{mukhoti2018evaluating,krishnan2020improving}, which point out that a well-calibrated model should be confident when its prediction is accurate and be uncertain when its prediction is inaccurate.

Generally, there are four possible outputs: (1) Accurate and Certain (AC), (2) Accurate and Uncertain (AU), (3) Inaccurate and Certain (IC), and (4) Inaccurate and Uncertain (IU).
We propose a calibration method encouraging the multi-view classifier to output more AC and IU samples. 
$y_{i}$ denotes the ground-truth label of sample $i$ while $\hat{y}_{i}$ denotes the prediction of model.
$\mathbf{\alpha}_{i}$ is the obtained Dirichlet parameter.
%while $\tilde{\mathbf{\alpha}}_{i}$ is the obtained Dirichlet parameter with the place of the correct class being set to $1$. 
$\tilde{\mathbf{\alpha}}_{i}=\mathbf{y}_{i}+\left(1-\mathbf{y}_{i}\right) \odot \mathbf{\alpha}_{i}$ is the Dirichlet parameters after removal of the correct evidence for the true class. 
Specifically, when the model makes a firm and accurate prediction ($\hat{y}_{i}=y_{i}$ and $\max(\mathbf{p}_{i}) \to 1$), we force it to give a relatively low uncertainty by increasing the total evidence strength ($S_i \to \infty$). 
When the model predicts falsely, we force it to give a high uncertainty by making misleading evidence shrink to zero ($\tilde{\mathbf{\alpha}}_{i} \to \mathbf{1}$):

\begin{equation}\label{eq_euc}
\begin{aligned}
\mathcal{L}_{E U C}(\mathbf{p}_i)= \lambda_e ( -\sum_{i \in\left\{\hat{y}_{i}=y_{i}\right\}} \max(\mathbf{p}_{i})\log \left(1-C/S_{i}\right)\\
+ \sum_{i \in\left\{\hat{y}_{i} \neq y_{i}\right\}}\text{KL}\left[D\left(\mathbf{p}_{i} \mid \tilde{\mathbf{\alpha}}_{i}\right) \| D\left(\mathbf{p}_{i} \mid \mathbf{1}\right)\right]),\\
 \lambda_e = \min(1.0, e/25),
 \end{aligned}
\end{equation}
where $C$ denotes the total number of event classes. $S_i$ represents the total Dirichlet strength and $C/S_i$ is the uncertainty of sample $i$. 
The KL term represents the Kullback-Leibler divergence between the wrong and uniform evidence distribution.  
$\it{D}(\cdot|\cdot)$ denotes the multinomial opinions formed by the Dirichlet parameter. 
The first term encourages AC outputs by ensuring more collected evidence, while the second term tries to give IU outputs by removing all wrong evidence $\tilde{\mathbf{\alpha}}_{i}$.
Meanwhile, considering that the learned evidence in the early epochs tends to be inaccurate, we also adopt an annealing coefficient $\lambda_e$ to dynamically adjust the weight of calibration loss. $e$ denotes the index of the current epoch. 
In the initial epoch, the class uncertainty used in the UCL loss is set to $1-\epsilon$, where $\epsilon$ is a very small value.

\textbf{Class Uncertainty}: 
Similar to the calculation of estimated class centroids, to make a trade-off between training stability and resource consumption, we update the class uncertainty in the full dataset every epoch and keep them fixed in memory for the duration of the whole epoch. 
The class uncertainty is the average uncertainty value of all the training samples within that class.

\subsection{Optimization Objective}
The optimization objective includes loss from the representation adjustment module and loss from the multi-view classifier module, termed as:
\begin{equation}\label{eq_total}
\begin{aligned}
\mathcal{L}_{Total} = \mathcal{L}_{Error}+\lambda_1\mathcal{L}_{EUC}+\lambda_2\mathcal{L}_{UCL}^v + \lambda_3\mathcal{L}_{Common} \\ v\in\{\textit{co-hashtag}, \textit{co-entity}, \textit{co-user}\},
\end{aligned}
\end{equation}
where $\lambda_1$, $\lambda_2$ and $\lambda_3$ are hyper-parameters. 
The latter two terms are in the representation module. 
Specifically, $\mathcal{L}_{UCL}^v$ is the proposed uncertainty-guided contrastive learning loss which aims to regularize representations in each view. 
$\mathcal{L}_{Common}$ is designed to tackle the deficiency of Dempster's rule in handling high-conflict data by ensuring the multi-view commonality. 
Here we denote the normalized embeddings over a batch of training samples from a specific view as $\mathbf{H}^v_{nor}, v\in\{\textit{co-hashtag}, \textit{co-entity}, \textit{co-user}\}$. 
The similarity of nodes $\mathbf{Sim}^{v}$ is computed as $\mathbf{H}^v_{nor} \cdot (\mathbf{H}^v_{nor})^{T}$.
$\mathcal{L}_{Common}$ gives the following constraint:
\begin{equation}\label{eq_common}
\begin{aligned}
    \mathcal{L}_{Common}=\left\|\mathbf{Sim}^{co-hashtag}-\mathbf{Sim}^{co-entity}\right\|_{F}^{2}\\+
    \left\|\mathbf{Sim}^{co-hashtag}-\mathbf{Sim}^{co-user}\right\|_{F}^{2}\\+
    \left\|\mathbf{Sim}^{co-entity}-\mathbf{Sim}^{co-user}\right\|_{F}^{2}.
\end{aligned}
\end{equation}
The former two terms $\mathcal{L}_{Error}$ and $\mathcal{L}_{EUC}$ are from the multi-view classifier module.
Specifically, $\mathcal{L}_{EUC}$ is proposed to calibrate the calculated multi-view uncertainty.
$\mathcal{L}_{Error}$ denotes the prediction error loss, integral to the classical cross-entropy loss function over the learned Dirichlet distribution.

\begin{equation}\label{eq_error}
\begin{aligned}
    \mathcal{L}_{Error}=\sum_{i}{\int\left[\sum_{j=1}^{C}-y_{i j} \log \left(p_{i j}\right)\right] \frac{1}{B\left(\mathbf{\alpha}_{i}\right)} \prod_{j=1}^{C} p_{i j}^{\mathbf{\alpha}_{i j}-1} d \mathbf{p}_{i}},
\end{aligned}
\end{equation}
where $\mathbf{y}_i$ is the true class distribution. $\mathbf{p}_i$ is the class assignment probabilities on a simplex and $\it{B}(\cdot)$ is the multinomial beta function.

\subsection{Time complexity analysis}\label{sec_time}
The total time complexity of UCL$_{SED}$ is about $O(\sum_{v\in V}N_e^v)$, where $V$ represents the set of views. $N_e^v$ denotes the total number of edges under the specific view $v$. That means the time complexity is approximately linear with
the multi-view graph size. Specifically, as node features are low-dimensional and $N_e^v\gg N$, the propagation of GNN encoder under all the views (Algorithm \ref{Alg_model} lines 8-9) takes $O(|V|Ndd'+ \sum_{v\in V}N_e^vd') = O(\sum_{v\in V}N_e^v)$, where $|V|$ denotes the total number of views. $N$ is the total number of messages. $d$ and $d'$ are the input and output dimensions of the propagation layer. The time complexity of UCL loss under all the views (Algorithm \ref{Alg_model} line 11) can be roughly estimated as $O(|V|NCd')$, where $C$ denotes the total number of classes. As for the EDL neural network (Algorithm \ref{Alg_model} line 12), its time complexity under all the views is about $O(|V|Nd'C)$. Additionally, it takes $O(|V|NC)$ to calculate the view-specific uncertainty (Algorithm \ref{Alg_model} line 13), $O(|V-1|N(C+1)^2)$ to multi-view uncertainty (Algorithm \ref{Alg_model} line 14), $O(NC)$ to $\mathcal{L}_{EUC}$ and $\mathcal{L}_{Error}$ (Algorithm \ref{Alg_model} line 15 and line 16), and $O(|V|(|V|-1)\sum_{b=1}^B|m_b|^2d')$ to $\mathcal{L}_{Common}$ (Algorithm \ref{Alg_model} line 17), where $|m_b|$ denotes the batch size and B is the number of mini-batches. Similar to UCL$_{SED}$, the total time complexity of UCL-EC$_{SED}$ is also about $O(\sum_{v\in V}N_e^v)$. The only difference lies in the additional $O(Nd')$ taken to calculate those prototypes, which can be neglected.

% The specific per-epoch training time in seconds of Events2012\_100, Events2018\_100, and CrisisLexT\_7 is 122.28, 109.23 and 3.31. Readers can refer to Sec.~\ref{sec_expset} for the detailed experimental setting.

\section{Experiments}\label{sec:exper}

\begin{figure*}
    \centering
    \includegraphics[width = 16.5cm]{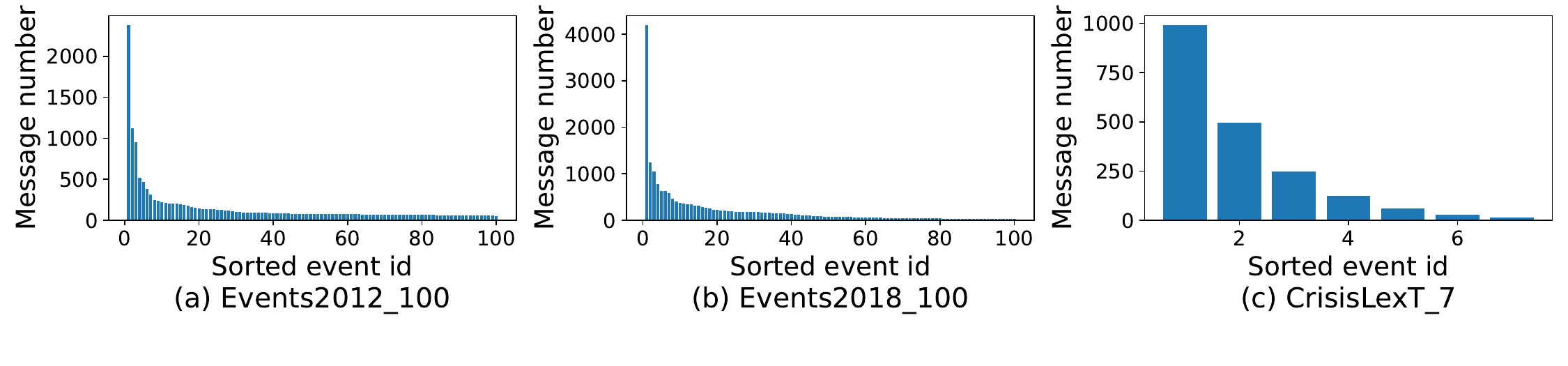}
    \caption{Detailed dataset statistics. (a), (b) and (c) show the number of messages per event on Events2012-100, Events2018-100, and CrisisLexT-7, respectively.}
    \label{fig_statistics}
\end{figure*}

\subsection{Experimental Setup}\label{sec_expset}
\subsubsection{Datasets and evaluation metric}
We conduct experiments on three imbalanced social event datasets: Events2012\_100, Events2018\_100, and CrisisLexT\_7. 
The former two datasets are sampled from Events2012~\cite{mcminn2013building} and Events2018~\cite{mazoyer2020french}, respectively. 
Considering that events in the original Events2012 and Events2018 datasets are with different frequencies, to reconstruct datasets following long-tail distribution, we select 100 events and reorder them based on the number of tweets each event contains. 
The final Events2012\_100 consists of 15019 tweets relating to 100 events. 
With maximally 2377 tweets and minimally 55 tweets per event, the imbalance ratio is about 43. 
Similarly, the final Events2018\_100 contains a total number of 19944 tweets. 
With maximally 4189 tweets and minimally 27 tweets per event, its imbalance ratio is about 155. 
CrisisLexT\_7 is sampled from CrisisLexT-26~\cite{olteanu2015expect}, a balanced dataset containing 26 crisis events in multiple languages. 
Here we only select crisis events in English. 
Meanwhile, to ensure the long-tail distribution of the reconstructed CrisisLexT\_7 dataset, we calculate the number of each event based on the exponential function: $n_i=n_{max}\gamma^i$, where $i$ is the event class index. 
$\gamma$ is set to $0.5$ in our experiments. 
With maximally 989 tweets and minimally 15 tweets per event, the imbalance ratio of CrisisLexT\_7 is about 66.
More details are shown in Fig.~\ref{fig_statistics}.

The statistics mentioned above depict the datasets for training.
Generally, an imbalanced training set, balanced validation, and test sets should be provided to obtain a more fair and accurate model evaluation for the long-tail recognition task. 
Thus, for the validation and test sets in our experiments, we select additional 20 and 30 tweets for each event, respectively. 
As for evaluation metrics, we simply choose the two commonly used metrics in classification tasks: Accuracy (ACC) and F1 value (F1).

\subsubsection{Our proposed algorithm}
According to how we learn and update the class prototypes, our uncertainty-guided class imbalance learning framework has the following variants:
(1) UCL$_{SED}$, which automatically learns the parameters of class prototypes and updates them every batch; 
(2) UCL-EC$_{SED}$, which uses the global class centroids as class prototypes and updates them every epoch. 
Please refer to Sec.~\ref{sec_ucl-ce} for more details.

\begin{table}
\caption{Dataset details under different views.}
\centering
\renewcommand\arraystretch{1.3}
\setlength{\tabcolsep}{1mm}
\begin{tabular}{ c |c| c|c}
\toprule
\multirow{2}{*}{View} &\multicolumn{3}{c}{Num of correct edges/Num of all edges}\\
\cline{2-4}
&Events2012\_100 &Events2018\_100 &CrisisLexT\_7  \\
\hline
\textit{co-hashtag} &0.7355&0.8572&0.8778\\
\textit{co-entity} &0.1976&0.6026&0.9257\\
\textit{co-user} &0.8847&0.7030&0.8707\\
\textit{all} &0.2323&0.7234&0.9121\\
\hline
% promotion &$\uparrow  3\%$ &$\uparrow 4\%$ &$\uparrow 5\%$ &$\uparrow 4\%$ \\
\bottomrule      
\end{tabular}  
\label{table_statistics}
\end{table}

\subsubsection{Baselines}
To verify the effectiveness of our proposed UCL$_{SED}$ and UCL-EC$_{SED}$ in detecting events from severely imbalanced datasets, we compare our methods with state-of-the-art techniques in the social event detection domain. 
Furthermore, to demonstrate the superiority of the proposed UCL loss in learning distinguishable representations for imbalanced datasets, we also compare our method with existing benchmark methods in long-tail recognition tasks. 
Overall, the selected baselines are listed in the following two groups.

\textbf{Social event detection methods:} 
The selected social event detection baselines are pre-trained language models: 
(1) Word2Vec~\cite{mikolov2013efficient} - we leverage the pre-trained word embeddings to get the message vectors and adopt a two-layer neural network to classify them; 
(2) BERT~\cite{devlin2018bert} - we finetune it on our datasets and make the final classification.
Topic models: 
(3) TwitterLDA~\cite{olteanu2015expect} obtains message representations by learning topic and word distributions. 
GNN-based models: 
(4) PP-GCN~\cite{peng2019fine}, which first builds a weighted adjacent matrix by measuring event similarity, then leverages a graph convolutional network trained by pair-wise sampling to obtain discriminate message representations; 
(5) KPGNN~\cite{cao2021knowledge}, which connects messages sharing common elements, then uses a multi-head graph attention network to learn message representations; 
(6) MVGAN~\cite{cui2021mvgan}, which learns message representations from both semantic and temporal views and uses an attention mechanism to fuse them; 
(7) ETGNN~\cite{ren2022evidential}, which learns message vectors from \textit{co-hashtag}, \textit{co-entity} and \textit{co-user} views and uses Dempster–Shafer theory to combine them.

\textbf{Long-tail recognition methods:} 
We also compare our methods with several long-tail recognition methods.
Note that 
(1) CE (i.e., Cross-Entropy) is the vanilla baseline, using the cross-entropy loss to train our multi-view framework. 
Other baselines include loss manipulation methods: 
(2) CB+Focal~\cite{cui2019class} (i.e., Class-Balanced Focal loss), which combines a re-weighting scheme with the original focal loss by assigning weights to different classes based on their sample numbers; 
(3) LDAM loss~\cite{cao2019learning}, which enforces class-dependent margins based on class frequencies; long-tail representation improvement methods: 
(4) Hybrid-PSC~\cite{wang2021contrastive}, which is a hybrid framework with a supervised contrastive learning branch for representation regularization and a classifier branch for bias elimination; 
(5) BCL~\cite{zhu2022balanced}, which further improves the original supervised contrastive learning by considering class averaging and class complement; 
(6) DRO-LT~\cite{samuel2021distributional}, which builds on distributional robustness optimization and explicitly seeks to improve the quality of representations for tail classes; 
(7) TSC~\cite{li2022targeted}, which uses pre-defined features to guide representation learning.

\subsubsection{Experimental Settings and Implementations}
The proposed UCL$_{SED}$ framework combines the backbone model ETGNN~\cite{ren2022evidential} with an additional representation adjustment module and a multi-view classifier module with an improved uncertainty calibration method. 
We set the batch size to $1500$, the layer of temporal-aware GNN to $2$, and the dimensions of the first and second GNN layers to $256$. 
As for the representation adjustment module, we set $\beta$ in the UCL and UCL-EC to $0.1$. 
Each EDL classifier is designed as a two-layer neural network with an activation layer ReLU in the multi-view classifier module. 
The hidden layer dimension of EDL is $128$. 
As for those hyper-parameters in the total optimization objective function, we set $\lambda_1$, which controls the intensities of uncertainty calibration, to 1, $\lambda_2$, which controls UCL to $0.1$, and $\lambda_3$ which ensures multi-view commonality to $0.5$.
The framework is trained using Adam optimizer with the learning rate $0.001$. The maximal training epoch is $100$.
Experiments are implemented in Python 3.8 and Pytorch 1.9 and conducted on $8\times$GeForce RTX 3090 GPU. To avoid the one-time occasionality, in comparison experiments, we perform 10 tests for all models and record the mean and standard deviation values.

\begin{table}
\caption{Comparison with social event detection methods.}
\centering
\renewcommand\arraystretch{1.3}
\setlength{\tabcolsep}{0.1mm}
\begin{tabular}{ c |c|c| c|c|c|c}
\toprule
\multirow{2}{*}{Methods} &\multicolumn{2}{c|}{Events2012\_100} &\multicolumn{2}{c|}{Events2018\_100} &\multicolumn{2}{c}{CrisisLexT\_7}  \\
\cline{2-7}
\multirow{2}{*}{}& ACC ($\%$) & F1 ($\%$) &ACC ($\%$)&F1 ($\%$)&ACC ($\%$)& F1 ($\%$)\\
\midrule
TwitterLDA~\cite{zhao2011comparing} &9.37±.44 &8.27±.49 &6.90±.51 &4.83±.60   &31.90±.53&22.59±.58  \\
Word2Vec~\cite{mikolov2013efficient} &74.67±.56 &74.89±.59 &35.17±.42 &33.89±.41  &44.29±.54 &37.80±.57  \\
BERT~\cite{devlin2018bert} &79.11±.38 &79.28±.46 & 56.39±.57 &54.07±.61   &69.43±.70 &66.29±.67 \\
PP-GCN~\cite{peng2019fine}   &63.33±.34  &54.62±.25 &70.00±.39 &50.99±.46   &73.33±.53&68.71±.44  \\
KPGNN~\cite{cao2021knowledge}    &73.33±.26  &59.08±.33 &76.67±.38 &61.90±.41 &75.67±.52&71.11±.60  \\
MVGNN~\cite{cui2021mvgan}    &  81.83±.29 &82.14±.31 &69.93±.42 &68.17±.47 &71.90±.49&70.55±.57  \\
ETGNN~\cite{ren2022evidential}    & 86.45±.23 &86.56±.27 &60.43±.31 &60.12±.33  & 76.67±.42 &74.30±.50 \\
% MVGAN~\cite{cao2021knowledge}    & .63±.01 &.59±.01 &.27±.03 &.75±.02 \\
\midrule
UCL$_{SED}$  & 92.21±.30 &92.01±.35  &78.16±.37&78.77±.41&80.81±.60 &80.67±.62 \\ 
UCL-EC$_{SED}$  &\textbf{93.18±.20} &\textbf{93.27±.29}  & \textbf{78.91±.25}&\textbf{79.11±.29}&\textbf{84.33±.41}&\textbf{83.96±53}   \\ 
\hline
% promotion &$\uparrow  3\%$ &$\uparrow 4\%$ &$\uparrow 5\%$ &$\uparrow 4\%$ \\
\bottomrule      
\end{tabular}  
\label{table_sed}
\end{table}

\subsection{Results and Comparisons}\label{sec:comparisons}
\subsubsection{Comparison with social event detection methods}
We first compare our approaches with $7$ competitive social event detection methods and report results in Table~\ref{table_sed}. 
By carefully analyzing the results, we have the following observations: 
(1) Our proposed model (i.e., UCL-EC$_{SED}$) achieves state-of-the-art performance on all three imbalanced datasets. 
On Events2018\_100, UCL-EC$_{SED}$ even surpasses ETGNN by about 19\%. 
Because our models capture the view-specific reliability better by adding the additional uncertainty calibration method and therefore, make up for the shortcomings of ETGNN in handling data whose most views are noisy.
% Besides, the great results of UCL$_{SED}$ and UCL-EC$_{SED}$ are also attributed to the fact that all the baselines neglect the severe class-imbalance distribution of the training data. 
% Thus, those baselines encounter a generalization problem during testing. 
% (2) Generally, the representations of GNN-based methods are superior to the pure pre-trained word or sentence embeddings. 
% Because text contents in social networks are often restricted to be pretty short and contain many informal expressions. 
% These characteristics cause the representations learned by pure text semantic mining not to be that distinguishable. 
(2) We also noticed that the results of those GNN-based baselines vary greatly on different datasets and ETGNN has difficulty in handling datasets whose views are noisy.
For example, on Events2012\_100, ETGNN achieves a remarkable accuracy improvement of 13.12\% compared with KPGNN. 
While on Events2018\_100, KPGNN has a relative improvement of 16.24\% compared with ETGNN. 
The connection qualities of the constructed social graphs determine this. 
Here we depict the detailed connection qualities in Table~\ref{table_statistics}.
Note that edges under the view ``\textit{all}'' are the union of edges under the above three single views. 
In KPGNN, information is propagated and aggregated over a homogeneous message graph constructed under the ``\textit{all}'' view. 
While in ETGNN, representations of three single views are learned independently, and the obtained view-specific results are further combined via Dempster-Shafer theory to get the final decision. 
As shown in Table~\ref{table_statistics}, on Events2012\_100, the connection quality of the \textit{co-entity} view is quite low, which also leads to the low quality of the ``\textit{all}'' view. 
Therefore, KPGNN performs badly. 
Meanwhile, considering most views (i.e., \textit{co-hashtag} and \textit{co-user}) are of relatively high quality, ETGNN can still obtain trusted results by combining multi-view results via Dempster-Shafer theory. 
However, on Events2018\_100, the connection qualities under most views (i.e., \textit{co-entity} and \textit{co-user}) are not that good. 
Meanwhile, the estimated uncertainty from ETGNN is not well calibrated. 
Therefore, ETGNN performs poorly.
To sum up, with the help of the UCL loss, our models are aware of the per-class representation status during training and make timely adjustments to their boundaries. 
More analysis of the UCL and UCL-EC losses will be made later.

\begin{figure*}
    \centering
    \includegraphics[width = 16.3cm]{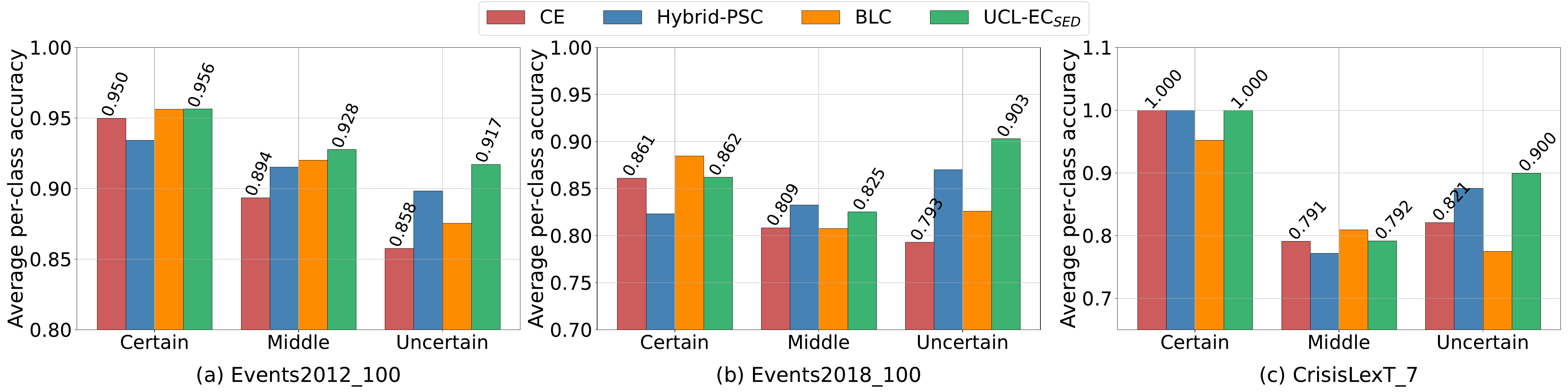}
    \caption{Average per-class accuracy and F1 value of the certain, middle, and uncertain groups.}
    \label{fig_group_result}
\end{figure*}

\begin{table}
\caption{Comparison with long-tail recognition methods.}
\centering
\renewcommand\arraystretch{1.3}
\setlength{\tabcolsep}{0.1mm}
\begin{tabular}{ c |c|c| c|c|c|c}
\toprule
\multirow{2}{*}{Methods} &\multicolumn{2}{c|}{Events2012\_100} &\multicolumn{2}{c|}{Events2018\_100} &\multicolumn{2}{c}{CrisisLexT\_7}  \\
\cline{2-7}
\multirow{2}{*}{}& ACC ($\%$) & F1 ($\%$) &ACC ($\%$)&F1 ($\%$)&ACC ($\%$)& F1 ($\%$)\\
\midrule
CE &85.77±.26 &85.88±.30 &73.73±.35 &74.00±.38   &74.76±.43 &72.31±.49  \\
CB+Focal~\cite{cui2019class} &87.67±.40 &86.84±.43 &75.57±.37 &75.23±.46  & 75.24±.53& 74.95±.66    \\
LDAM~\cite{cao2019learning} &89.83±.29 &89.95±.24&77.30±.36 &78.14±.44  &76.84±.49&76.61±.42     \\
% SSP~\cite{} & & & &   & &  \\
Hybrid-PSC~\cite{wang2021contrastive} &88.57±.26 &88.65±.33 &76.87±.39  &76.60±.47   &78.10±.43 &76.20±.52  \\
BCL~\cite{zhu2022balanced} &90.83±.34 &90.98±.37 &77.33±.24 &78.16±.26 &81.90±.47 &81.32±.45    \\
DRO-LT~\cite{samuel2021distributional} &89.43±.28 &89.42±.25 &77.07±.34&77.52±.40 &78.86±.45 &78.30±.53   \\
TSC~\cite{li2022targeted} &90.33±.24 & 90.80±.26 &77.40±.35 &78.33±.36   &81.43±.42 &80.58±.48  \\
\midrule
UCL$_{SED}$  & 92.21±.30 &92.01±.35  &78.16±.37&78.77±.41&80.81±.60 &80.67±.62 \\ 
+CB &92.83±.35 &92.80±.38  &78.70±.45 &78.97±.50 &81.57±.68 &81.25±.74 \\ 
\midrule
UCL-EC$_{SED}$  &93.18±.20&93.27±.29  &78.91±.25&79.11±.29&84.33±.41&83.96±.53  \\  
+CB &\textbf{93.67±.31} &\textbf{93.66±.36}  &\textbf{79.33±.42} &\textbf{79.27±.41} &\textbf{84.60±.51} &\textbf{84.24±.58} \\ 
\hline
% promotion &$\uparrow  3\%$ &$\uparrow 4\%$ &$\uparrow 5\%$ &$\uparrow 4\%$ \\
\bottomrule      
\end{tabular}  
\label{table_ltr}
\end{table}

\subsubsection{Comparison with long-tail recognition methods}\label{sec_long_tail}
Note that this work focuses on social event detection in imbalanced data.
The proposed UCL method aims to enhance the generalization capacity by regularizing the representation learning. 
Thus, we also compare our methods with the vanilla CE baseline and six state-of-the-art long-tail recognition methods, especially with those long-tail representation improvement methods. 
The results are presented in Table~\ref{table_ltr}. 
Our UCL-EC$_{SED}$ consistently outperforms all the baselines on all three datasets, emphasizing the superiority of enforcing margins in the feature space under the guidance of per-class uncertainty.

We here analyze the experimental results in detail. 
As seen from Table~\ref{table_ltr}, CE performs worst among all the baselines. 
This is due to the limitation of the original cross-entropy loss in handling imbalanced data.
To deal with the imbalanced data distribution, CB+Focal manipulates the focal loss by assigning more weights to classes with less effective samples. 
As a loss re-weighting method, CB+Focal slightly improves the detection results. 
LDAM is also a loss manipulation method. 
Instead of modifying per-class weight, it sets different logit margins for different classes based on frequency. 
LDAM obtains a relatively good performance. 
This method focuses on improving the output layer of the final classifier while the remaining methods consider improving the output layer of the GNN model. 
In Table~\ref{table_ltr}, Hybrid-PSC, which applies prototypical supervised contrastive loss to learn distinguishable features, outperforms the CE counterpart. 
For example, on ACC, Hybrid-PSC outperforms CE by 2.8\%, 3.14\%, and 3.34\% on Events2012\_100, Events2018\_100, and CrisisLexT\_7, respectively. 
This validates the idea that better representation can help distinguish event classes. 
We also noticed that other methods that tailor the contrastive learning loss for imbalanced datasets achieve better results than Hybrid-PSC. 
For example, TSC pre-computes a set of targets uniformly to ensure data balance. 
However, it is not flexible enough as it has no ability to make proper adjustments for different classes.
DRO-LT extends prototypical contrastive learning by introducing distributional robustness. 
It learns separable per-class representation by pushing and pulling towards a worst-case possible distribution. 
BCL is also a representation improvement method that applies contrastive learning. 
It implements class-averaging and class-complement to the original contrastive learning loss to enhance representation learning. 
Table~\ref{table_ltr} shows a significant performance boost when comparing the tailored BCL with Hybrid-PSC. 
DRO-LT and BCL methods are tailored for imbalanced data under frequency guidance. 
However, as observed from Fig~\ref{fig_toy}, evidential uncertainty is a better indicator of model generalization capacity than class frequency.
We, therefore, enforce larger margins for uncertain classes during representation learning. 
The result that our UCL-EC$_{SED}$ surpasses all the baselines validates the superiority of our uncertainty-guided learning.

For a more fine-grained understanding, we also split all the labels into three groups based on their measured per-class uncertainties and plot the final group results in Fig.~\ref{fig_group_result}. 
Concretely, we divide all uncertainty values into three intervals. 
Assume the maximal and minimal class uncertainty values are denoted as $U_{max}$ and $U_{min}$, classes whose uncertainty values are within $[U_{min}, U_{min}+1/3(U_{max}-U_{min})]$ are split into certain classes.
Classes within $[U_{min}+1/3(U_{max}-U_{min}), U_{min}+2/3(U_{max}-U_{min})]$ are middle and the rest are uncertain.
Fig.~\ref{fig_group_result} reveals that our UCL-EC$_{SED}$ achieves a large gain on the uncertain group without sacrificing the detection performances on the certain and middle groups. 
This further demonstrates the robustness of our model. 
Adding proper uncertainty-guided margins during training makes class representations in all groups more separable.
Besides, it is observed that UCL-EC$_{SED}$ performs even better than UCL$_{SED}$. We argue this is due to the gap in sample distribution in different batches. 
% In UCL$_{SED}$, the learned prototypes get updated every batch and, therefore,  fluctuate greatly. 
% The model may encounter a training instability challenge.

\begin{figure*}
    \centering
    \includegraphics[width = 16.3cm]{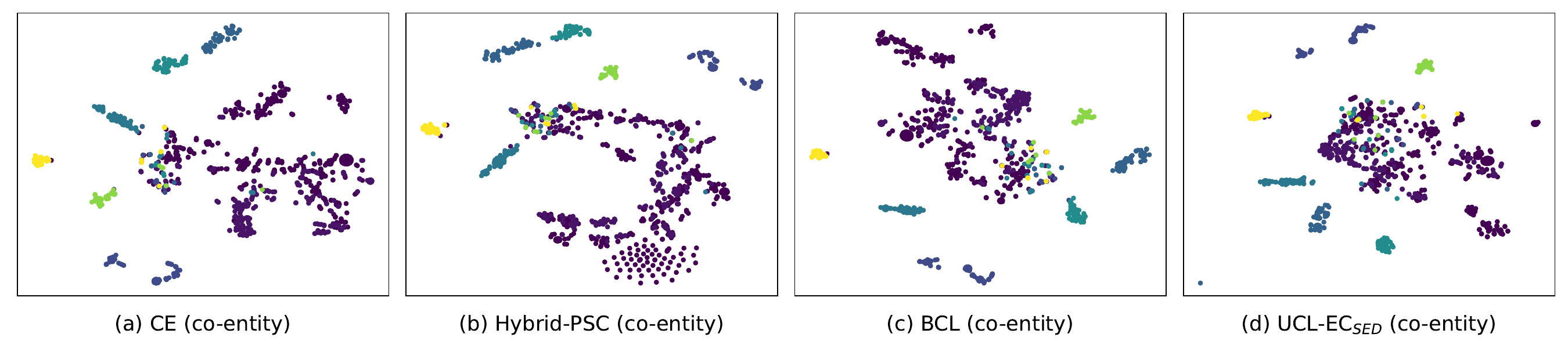}
    \caption{t-SNE visualization of the learned features of \textit{co-entity} view on Events2012\_100. Here we randomly select $8$ events which are drawn in different colors.
     }
    \label{fig_tsne}
\end{figure*}

\begin{figure*}
    \centering
    \includegraphics[width = 16.3cm]{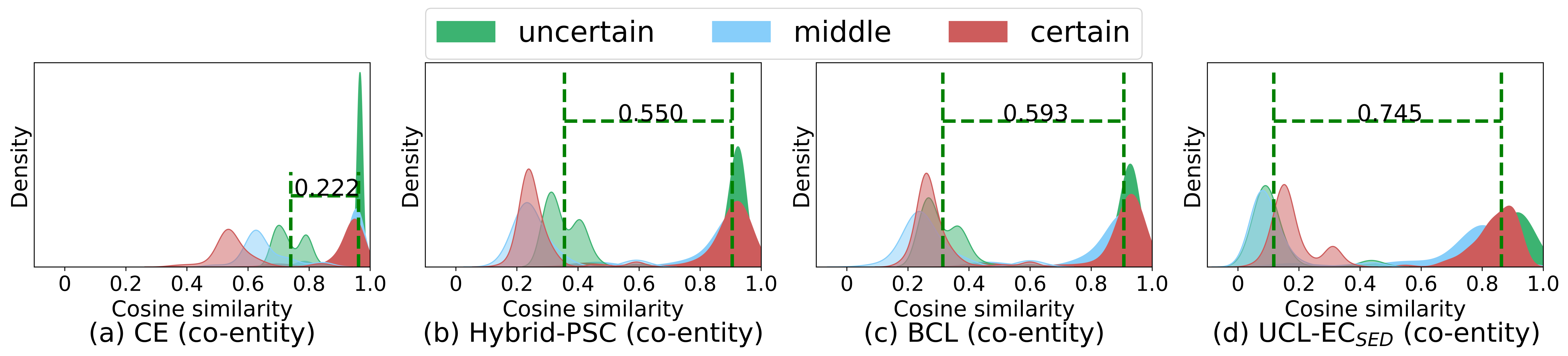}
    \caption{Visualization of mean intra-class and inter-class cosine similarity distribution of \textit{co-entity} view on Events2012\_100. 
    % The first, second, and third rows show the similarity distribution of \textit{co-hashtag}, \textit{co-entity}, and \textit{co-user} views, respectively. 
    Dark color area indicates intra-class similarities while light color area indicates inter-class ones. 
    Uncertain, middle, and certain classes are plotted separately. 
    Average inter-class cosine similarity and average intra-class similarity of the uncertain group are marked with dashed green vertical lines. }
    \label{fig_visualize}
\end{figure*}

Remark: unlike most classical long-tail recognition methods that directly act on the classifier (e.g., re-weighting strategies), our work tends to solve the imbalance problem by manipulating the latent feature space. 
By ensuring the learned representations of minority event classes are well-separated from other event classes, it becomes much easier to recognize them. 
Our methods work differently from classical methods, which means they are parallel and may complement each other. 
To validate this opinion, we here further incorporate the re-weighting strategy (i.e., CB in \cite{cui2019class}) into the classification Error loss $\mathcal{L}_{Error}$ of our framework and record the results. 
Encouragingly, the performance gets further improved.
For example, UCL-EC$_{SED}$ + CB gets a further $0.49\%$ improvement on ACC on Events2012\_100.

\subsection{Representation Visualization}\label{sec:Visualization}\label{sec_visualization}
To make a better analysis of representations learned by baselines (CE, Hybrid-PSC, BCL) and our model (UCL-EC$_{SED}$), in Fig.~\ref{fig_tsne}, we plot the t-SNE results of eight randomly selected events of \textit{co-entity} view on Events2012\_100. Obviously, the boundary learned by our model is less blurry compared to other baselines. Compared to Hybrid-PSC which adopts the original PSC, the class overlapping problem gets well alleviated by our work. To further demonstrate the extent to which our UCL loss helps adjust a clear separable boundary in the latent space, we also visualize the mean intra-class similarity distribution and mean inter-class similarity distribution of samples in certain, middle and uncertain groups on the Events2012\_100 dataset. 
The results are plotted in Fig.~\ref{fig_visualize}. 
Note that the dark area represents intra-class cosine similarities while the light color represents inter-class ones. 
As can be observed, the representations of UCL-EC$_{SED}$ are the best for all three groups (uncertain, middle, and certain) under all three views. 
The inter-class similarity of UCL-EC$_{SED}$ gets lower compared to the competitive baseline - BCL, which is attributed to the added uncertainty-guided margin in the UCL loss. 
By adding a tunable margin, UCL pushes the distribution of other classes away from uncertain classes and, therefore, gets more separable representations.
Consistent with the results in Table~\ref{table_ltr}, representations learned by CE are the worst. 
Their intra-class similarities and inter-class similarities under all three views are closest. 
Compared to CE, Hybrid-PSC decreases the inter-class similarities significantly, owing to the ability of contrastive learning to push inter-class samples away.
Similar to Hybrid-PSC, BCL decreases inter-class similarities, especially for uncertain samples. 
We argue that may be because tail classes are more likely to be uncertain classes. 
BCL has a stronger ability to deal with tail classes thanks to its class-averaging and class-compensation strategies. Thus BCL also performs well in uncertain classes.

\subsection{Uncertainty Analysis}\label{sec:}
\begin{figure*}
    \centering
    \includegraphics[width = 16.3cm]{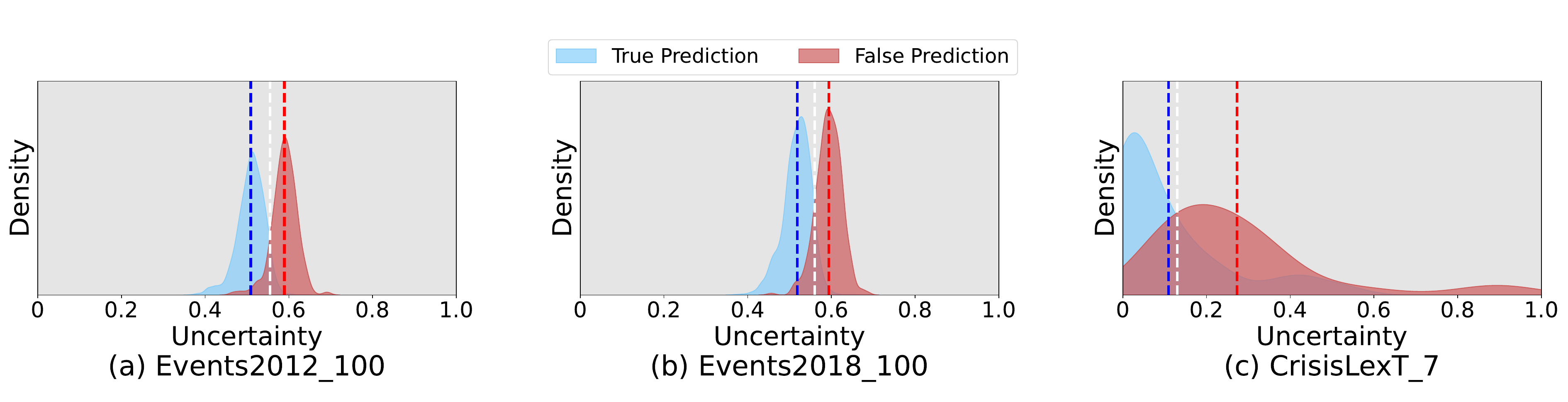}
    \caption{Visualization of uncertainty distribution on Events2012\_100, Events2018\_100, and CrisisLexT\_7.}
    \label{fig_uncertainty}
\end{figure*}

In this section, we conduct experiments to validate that the estimated uncertainty is highly correlated to the model performance and therefore, is a good indicator to adjust representation distribution. 
%We also demonstrate that our model, which combines multi-view results based on uncertainty, is robust to noisy data. 
We here visualize the estimated uncertainty of the true and false predictions in the validation set. 
The results are drawn in Fig.~\ref{fig_uncertainty}.
Obviously, on all three datasets, higher uncertainties are usually estimated for those false predictions while lower uncertainties are more likely to belong to those true predictions. 
This observation implies the effectiveness of using estimated uncertainty to indicate the status of representation learning since the estimated uncertainty is correlated to the prediction performance.

\begin{figure*}
    \centering
    \includegraphics[width = 16.3cm]{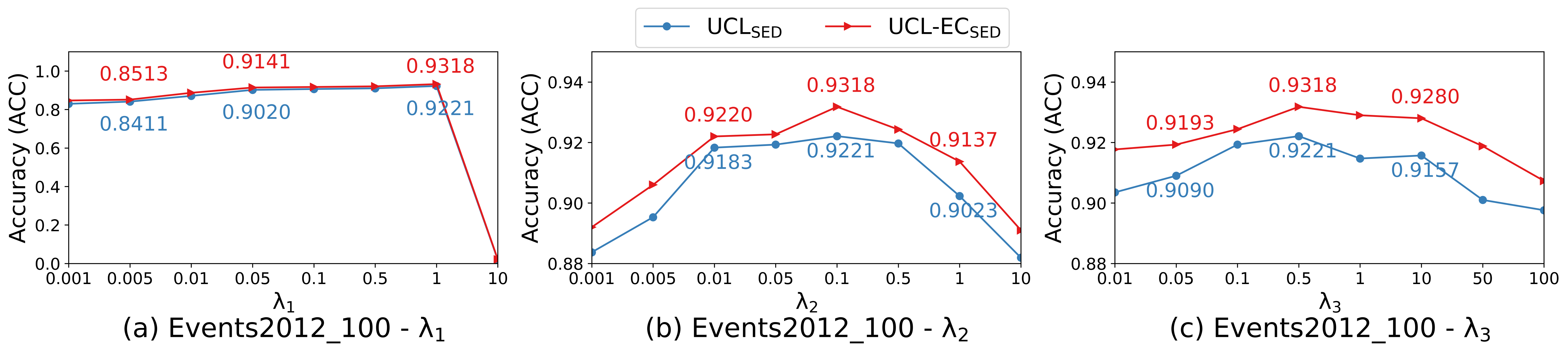}
    \caption{Hyper-parameter sensitivity analysis.}
    \label{fig_hyper}
\end{figure*}

\subsection{Hyper-parameter Sensitivity}\label{sec:}
In this section, we study the sensitivities of parameters $\lambda_1$, $\lambda_2$ and $\lambda_3$ in the optimization objective function (i.e., Eq.~\ref{eq_total}).
Due to the page limit, we only plot the results on Events2012\_100 in Fig.~\ref{fig_hyper}. 

\subsubsection{Analysis of coefficient $\lambda_1$.} 
The hyper-parameter $\lambda_1$ in Eq.~\ref{eq_total} controls how accurate the estimated uncertainty is. We vary it from $0.001$ to $10$. 
The results are shown in Fig.~\ref{fig_hyper}(a). 
With the increase of $\lambda_1$, the performance rises first. 
Because the uncertainty calibration loss helps in both classification and representation adjustment modules. 
By forcing wrong evidence to shrink to zero and highlighting those correct parts, it assists the learning of EDL neural networks. 
Meanwhile, more accurate uncertainty estimation helps adjust representation. 
However, the result drops rapidly when $\lambda_1$ reaches a relatively large value. 
Because too much emphasis is placed on eliminating wrong evidence in early training. 
In early epochs, misclassified samples are dominant. 
A large $\lambda_1$ may cause premature convergence to the uniform distribution, thus preventing the model from classifying correctly.

\subsubsection{Analysis of coefficient $\lambda_2$.} 
The hyper-parameter $\lambda_2$ in Eq.~\ref{eq_total} controls the impact of the representation adjustment module, which intends for the GNN model to learn separable features. 
We vary it from $0.001$ to $10$. 
The results are shown in Fig.~\ref{fig_hyper}(b). 
Similarly, as $\lambda_2$ becomes larger, the accuracy scores increase first and then decrease. 
Because in this work, instead of measuring the quality of the learned representations, the selected metrics in fact focus on the classification results. 
Though better representation helps better classification. 
The overwhelming emphasis on representation may lead to the under-learning of classifiers.

\subsubsection{Analysis of coefficient $\lambda_3$.} 
The hyper-parameter $\lambda_3$ in Eq.~\ref{eq_total} controls the consistency constraint on the three similarity matrices under different views. 
We vary it from $0.01$ to $100$. 
As shown in Fig.~\ref{fig_hyper}(c), the performance of our proposed framework is not very sensitive to $\lambda_3$ when it is within a reasonable range.

\subsection{Ablation Study}
In this section, we conduct quantitative ablation studies to analyze the components of our model. 
We record the results on Event2012\_100 in Table~\ref{table_ablation} for illustration purposes. 
% Concretely, we explore whether the UCL loss or the UCL-EC loss is helpful for imbalanced social event detection. 
% We also study the advantages of using the uncertainty calibration method. 

\subsubsection{Ablation study on the UCL and UCL-EC loss.}\label{sec:}
We argue that our work's proposed uncertainty-guided contrastive learning losses (i.e., UCL and UCL-EC) are expected to learn better features for imbalanced datasets, leading to better detection performance. 
To verify it, in the two proposed models UCL$_{SED}$ and UCL-EC$_{SED}$, we replace UCL and UCL-EC with their vanilla versions which remove the uncertainty-guided margins. 
As shown in Table~\ref{table_ablation}, without the uncertainty-guided margin, the results have a drop. 
To make a more careful comparison, we also add a fixed margin (i.e., the $+m$ strategy) to PSC and PSC-EC and report the results. 
When adding a fixed margin, the results are slightly better than those of the vanilla version losses but worse than uncertainty-guided ones. 
This further demonstrates the superiority of UCL and UCL-EC in learning separable representation.
What's more, to demonstrate the superiority of our uncertainty-related approach,
we also compare our methods with the baseline adopting the $+dm$ strategy but removing the uncertainty-related parts (i.e., the calibration part in the classification head and the uncertainty adjustment part in the representation head) in the whole framework. 
In this baseline, the per-class error rate of the training set is used to adjust representation. 
As demonstrated in Table~\ref{table_ablation}, it gets a great result while still worse than our uncertainty approach.
Furthermore, in comparison to this baseline, our framework offers enhanced interpretability and reliability. It achieves this by providing uncertainty values during the prediction process.

\begin{table}
\caption{Ablation studies of proposed models on Events2012\_100. 
The check mark indicates which losses are applied in the framework. 
$\mathcal{L}_{PSC}^v$ denotes the original Prototypical Supervised Contrastive loss. 
$+m$ means adding a fixed margin in the original PSC loss (introduced in Sec.~\ref{sec_psc} and Sec.~\ref{sec_psc_ucl}). 
Similarly, $+dm$ means adding a dynamic margin controlled by the per-class error rate of the training set in each epoch. 
The superscript $v$ (i.e., $v$ in $\mathcal{L}_{UCL}^v$, $\mathcal{L}_{PSC}^v$, $\mathcal{L}_{UCL-EC}^v$ and $\mathcal{L}_{PSC-EC}^v$) denotes the three views, $v\in\{\textit{co-hashtag}, \textit{co-entity}, \textit{co-user}\}$. }
\centering
\renewcommand\arraystretch{1.2}
\setlength{\tabcolsep}{0.5mm}
\begin{tabular}{ c |c|ccc|c|cc}
\toprule
Methods&  $\mathcal{L}_{UCL}^v$&  $\mathcal{L}_{PSC}^v$& $+m$ &$+dm$&$\mathcal{L}_{EUC}$& ACC & F1  \\
\midrule
UCL$_{SED}$  &\Checkmark & & &&\Checkmark  &0.9221 &0.9201   \\ 
UCL$_{SED}$  &  &\Checkmark  &  &&\Checkmark  &0.8825  &0.8838   \\ 
UCL$_{SED}$ & &\Checkmark&\Checkmark & &\Checkmark      &0.8957 &0.8955   \\ 
UCL$_{SED}$ & &\Checkmark& &\Checkmark &      &0.9073 &0.9056   \\ 
UCL$_{SED}$ &\Checkmark  &  &  & & &0.8193 &0.8179   \\ 
\midrule
\midrule
&  $\mathcal{L}_{UCL-EC}^v$&  $\mathcal{L}_{PSC-EC}^v$& $+m$&$+dm$&$\mathcal{L}_{EUC}$& ACC & F1  \\
\midrule
UCL-EC$_{SED}$ &\Checkmark &  & &&\Checkmark  &0.9318 &0.9327   \\ 
UCL-EC$_{SED}$   &  &\Checkmark  & & &\Checkmark   &0.8910 &0.8900   \\ 
UCL-EC$_{SED}$ & &\Checkmark&\Checkmark &&\Checkmark      &0.8983 &0.8975   \\ 
UCL-EC$_{SED}$ & &\Checkmark& &\Checkmark &     &0.9183 &0.9145   \\ 
UCL-EC$_{SED}$ &\Checkmark  &  &  & & &0.8230 &0.8300  \\ 
\hline
% promotion &$\uparrow  3\%$ &$\uparrow 4\%$ &$\uparrow 5\%$ &$\uparrow 4\%$ \\
\bottomrule      
\end{tabular}  
\label{table_ablation}
\end{table}

\subsubsection{Ablation study on the uncertainty calibration method}
The uncertainty calibration method (i.e., $\mathcal{L}_{EUC}$) is quite important to our model.
As can be seen in Table~\ref{table_ablation}, if we remove $\mathcal{L}_{EUC}$, the detection results have a significant decrease. For example, on ACC, the result of UCL$_{SED}$ has a 10.28\% drop.
We argue that uncertainty is important in the representation adjustment and multi-view classifier modules. 
We use per-class uncertainty in the representation module to set a margin for each class. 
In the classifier module, we utilize the view-specific uncertainty of each sample to make the multi-view combination and obtain the final result. 
Thus, we need the estimated uncertainty to be as accurate as possible, which makes the uncertainty calibration method indispensable.

\begin{table}
\caption{Time consumption. The table records the per-epoch running time of model training in seconds.}
\centering
\renewcommand\arraystretch{1.2}
\setlength{\tabcolsep}{1.2mm}
\begin{tabular}{ c ||c|c||c|c}
\toprule
  Dataset&\multirow{4}{*}{UCL$_{SED}$}& Time&\multirow{4}{*}{UCL-EC$_{SED}$} & Time \\
\cline{0-0} 
\cline{3-3}
\cline{5-5}
Events2012\_100& &122.28  & &135.43   \\
Events2018\_100& &109.23 & &111.77   \\
CrisisLexT\_7&  &3.31 & &3.35   \\
 
\hline
% promotion &$\uparrow  3\%$ &$\uparrow 4\%$ &$\uparrow 5\%$ &$\uparrow 4\%$ \\
\bottomrule      
\end{tabular}  
\label{table_time}
\end{table}
\subsection{Time consumption}
We record the time consumption information of UCL$_{SED}$ and UCL-EC$_{SED}$ in Table \ref{table_time}. As can be observed, overall, the per-epoch training time of UCL$_{SED}$ and UCL-EC$_{SED}$ is comparable, with UCL$_{SED}$ taking slightly less time than the latter. This is consistent with the time complexity analysis in Sec.~\ref{sec_time}. Compared to UCL$_{SED}$, in each epoch, UCL-EC$_{SED}$ needs extra calculation to update those prototypes. 
\section{Conclusion and future work}\label{sec:conclusion}
This paper proposes a novel uncertainty-guided class imbalance learning framework, namely UCL$_{SED}$, and its variant - UCL-EC$_{SED}$, for imbalanced SED tasks.
As a label-dependent representation regularization technique, the UCL$_{SED}$ aims to improve the model generalization capability by enhancing representation learning for all classes, especially for those uncertain ones. 
Specifically, we design a novel uncertainty-guided contrastive learning loss that assigns larger margins for those more uncertain classes to manipulate separable representation boundaries. 
Meanwhile, we propose a multi-view combination architecture with an additional calibration method to ensure accurate and robust uncertainty estimation.
The final detection result is combined via Dempster-Shafer theory under the supervision of uncertainty calibration. 
Experimental results verify the superiority of our model. 

However, there are also some limitations. 
Both UCL$_{SED}$ and UCL-EC$_{SED}$ learn only one single prototype for each class, which makes them insufficient to handle complicated classes that follow a multimodal distribution. 
We leave the extension to multiple prototypes as future work.

\section*{Acknowledgement}
This work is supported by National Key R\&D Program of China through grant 2022YFB3104700, NSFC through grants 62322202, U21B2027, 62002007, 61972186 and 62266028, Yunnan Provincial Major Science and Technology Special Plan Projects through grants 202302AD080003, 202202AD080003 and 202303AP140008, General Projects of  Basic Research in Yunnan Province through grant 202301AS070047, 202301AT070471, Beijing Natural Science Foundation through grant 4222030, S\&T Program of Hebei through grants 20310101D and 21340301D, and the Fundamental Research Funds for the Central Universities. 
Jia Wu was supported by the Australian Research Council (ARC) Project DP230100899.
Philip S. Yu was supported in part by NSF under grants III-1763325, III-1909323, III-2106758, and SaTC-1930941.

\bibliography{ref}
%======================
%===========================================================================

\begin{IEEEbiography}
[{\includegraphics[width=1in,height=1.1in,clip,keepaspectratio]{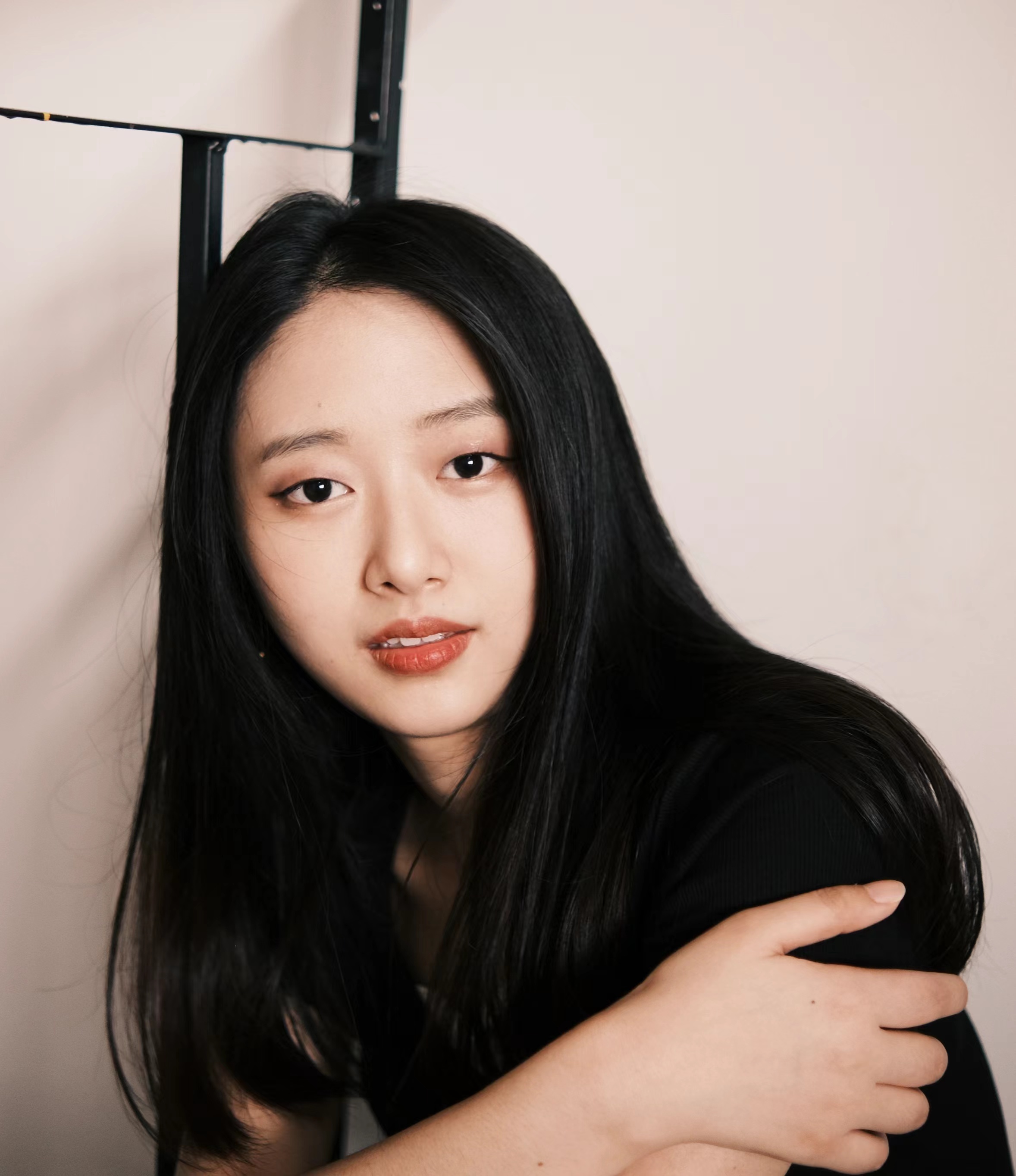}}]
{Jiaqian Ren} is currently a Ph.D. candidate in Institute of Information Engineering, Chinese Academy of Sciences. 
Her research interests include social event mining and graph representation learning.
\end{IEEEbiography}

\begin{IEEEbiography}
[{\includegraphics[width=1in,height=1.25in,clip,keepaspectratio]{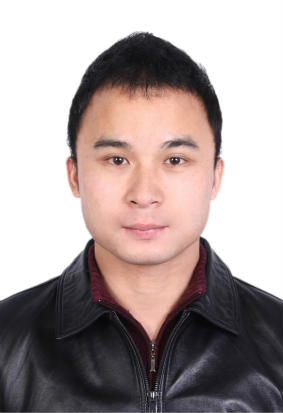}}]
{Hao Peng} is currently an Associate Professor at the School of Cyber Science and Technology in Beihang University. 
His current research interests include data mining, machine learning, and deep learning.
To date, Dr Peng has published over 100+ research papers in top-tier journals and conferences, including the IEEE TPAMI, TKDE, TC, ACM TOIS, TKDD, and Web Conference. 
He is the Associate Editor of International Journal of Machine Learning and Cybernetics (IJMLC).
\end{IEEEbiography}

\begin{IEEEbiography}
[{\includegraphics[width=1in,height=1.25in,clip,keepaspectratio]{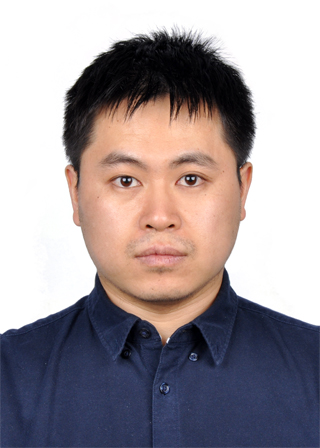}}]
{Lei Jiang} is an associate professor in Institute of Information Engineering, Chinese Academy of Sciences. 
His current research interest include network security and social computing.
\end{IEEEbiography}

\begin{IEEEbiography}
[{\includegraphics[width=1in,height=1.25in,clip,keepaspectratio]{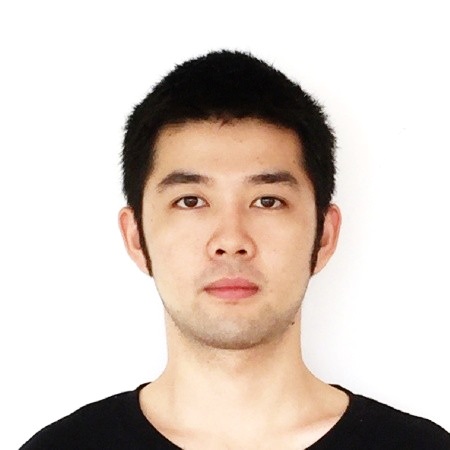}}]
{Zhiwei Liu} is currently a Research Scientist at Salesforce AI Research. His research interests include graph representation learning, recommender system and natural language understanding. 
He has published over 30 original research works in top-tier journals and conferences, including ACM TIST, Web Conference, SIGIR, CIKM, WSDM, and EMNLP.
\end{IEEEbiography}

\begin{IEEEbiography}
[{\includegraphics[width=1in,height=1.1in,clip,keepaspectratio]{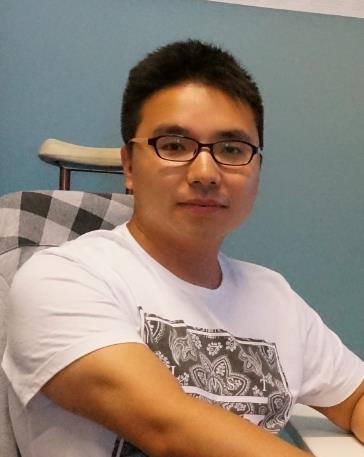}}]
{Jia Wu} is currently the Research Director for the AI-enabled Processes (AIP) Research Centre and an ARC DECRA Fellow in the School of Computing, Macquarie University, Sydney, Australia. 
His current research interests include data mining and machine learning. 
Since 2009, he has published 100+ referred journal and conference papers, including TPAMI, TKDE, TNNLS, TMM, TKDD, NIPS, WWW, and KDD. 
Dr. Wu was the recipient of SDM’18 Best Paper Award in Data Science Track, IJCNN’17 Best Student Paper Award, and ICDM’14 Best Paper Candidate Award. 
He is the Associate Editor of the ACM Transactions on Knowledge Discovery from Data (TKDD) and Neural Networks (NN). 
\end{IEEEbiography}

\begin{IEEEbiography}
[{\includegraphics[width=1in,height=1.1in,clip,keepaspectratio]{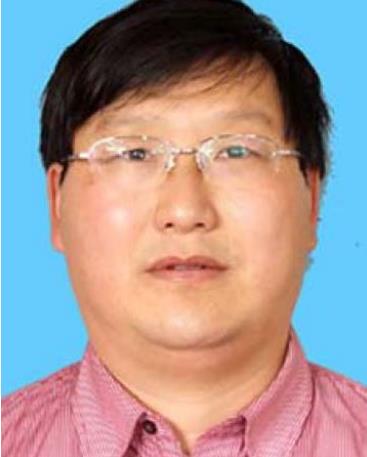}}]
{Zhengtao Yu} received the Ph.D. degree in computer application technology from the Beijing Institute of Technology, Beijing, China, in 2005.
He is currently a Professor with the School of Information Engineering and Automation, Kunming University of Science and Technology, China. His current research interests include natural language process, image processing, and machine learning. 
\end{IEEEbiography}

\begin{IEEEbiography}[{\includegraphics[width=1in,height=1.25in,clip,keepaspectratio]{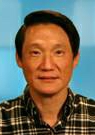}}]{Philip S. Yu} is a Distinguished Professor and the Wexler Chair in Information Technology at the Department of Computer Science, University of Illinois at Chicago. 
Before joining UIC, he was at the IBM Watson Research Center, where he built a world-renowned data mining and database department. 
He is a Fellow of the ACM and IEEE. 
Dr. Yu has published more than 1,200 referred conference and journal papers cited more than 179,000 times with an H-index of 189. 
Dr. Yu was the Editor-in-Chiefs of ACM Transactions on Knowledge Discovery from Data (2011-2017) and IEEE Transactions on Knowledge and Data Engineering (2001-2004). 
\end{IEEEbiography}

\end{document}